\documentclass[lettersize,journal]{IEEEtran}
\usepackage{amsmath,amsfonts}
\usepackage{algorithmic}
\usepackage{algorithm}
\usepackage{array}
\usepackage[caption=false,font=normalsize,labelfont=sf,textfont=sf]{subfig}
\usepackage{textcomp}
\usepackage{stfloats}
\usepackage{url}
\usepackage{verbatim}
\usepackage{graphicx}
\usepackage{cite}
\usepackage{multirow}
\usepackage{booktabs}
\usepackage[table]{xcolor}
\usepackage{diagbox}
\usepackage{colortbl} % for rowcolors
\usepackage{siunitx} 
\usepackage{diagbox}
\usepackage{pifont}    % 提供 \XSolidBrush
\usepackage{bbding}    % 提供 \CheckmarkBold

\definecolor{darkgreen}{RGB}{0,100,0}
\begin{document}

\title{FVeinSyn: Synthetic Finger Vein Image Generator}
\author{Yifan Wang, Jie Gui, Adams Wai Kin Kong, Baosheng Yu, Changsheng Chen, Qi Li, Zhenan Sun, James Tin-Yau Kwok, and Alex Kot
	% <-this % stops a space
	\thanks{This work was supported in part by the grant of the National Science Foundation of China under Grant 62172090; the CIPS-SMP-Zhipu Large Model Fund; the China Scholarship
		Council under Grant 202506090188. We thank the Big Data Computing Center of Southeast University for providing the facility support on the numerical calculations. (Corresponding author: Jie Gui.)}
	\thanks{Y. Wang and J. Gui are with the School of Cyber Science and Engineering, Southeast
		University, Nanjing 210000, China. J. Gui is also with Engineering Research Center of Blockchain Application, Supervision and Management (Southeast University), Ministry of Education; Purple Mountain Laboratories, Nanjing 210000, China (e-mail: 230239767@seu.edu.cn, guijie@seu.edu.cn).}
    \thanks{A. Kong is with the School of Computer Science and Engineering, Nanyang Technological University, Singapore 639798. (e-mail:adamskon@ntu.edu.sg).}
    \thanks{B. Yu is with the Lee Kong Chian School of Medicine, Nanyang Technological University, Singapore 639798 (e-mail: baosheng.yu@ntu.edu.sg).}
    \thanks{C. Chen and A. Kot are with Faculty of Engineering, Shenzhen MSU-BIT University, Shenzhen 518000, China. A. Kot is also with VinUniversity, Hanoi, Vietnam 100000, and also with Nanyang Technological University, Singapore 639798. (e-mail: cschen@szu.edu.cn; EACKOT@ntu.edu.sg).}
    \thanks{Q. Li and Z. Sun are with the New Laboratory of Pattern Recognition and the State Key Laboratory of Multimodal Artificial Intelligence Systems, Institute of Automation, Chinese Academy of Sciences, Beijing 100190, China, and also with the School of Artificial Intelligence, University of Chinese Academy of Sciences, Beijing 100049, China (e-mail: qli@nlpr.ia.ac.cn; znsun@nlpr.ia.ac.cn).}
\thanks{J. Kwok is with the Department of Computer Science and Engineering, The Hong Kong University of Science and Technology, Hong
		Kong 999077, China (e-mail: jamesk@cse.ust.hk).}
		\thanks{This paper has been accepted by IEEE Transactions on Pattern Analysis and Machine Intelligence.}
	}
\markboth{Journal of \LaTeX\ Class Files,~Vol.~14, No.~8, August~2021}%
{Shell \MakeLowercase{\textit{et al.}}: A Sample Article Using IEEEtran.cls for IEEE Journals}


\maketitle

\begin{abstract}
A major challenge in finger vein recognition is the lack of large-scale public datasets. Existing datasets contain few identities and limited samples per finger, restricting the advancement of deep learning-based methods. To address this, we propose FVeinSyn, a large-scale controllable synthetic data generation framework for finger vein. It explicitly decouples synthesis of vascular topology and imaging appearance to mitigate the limitations caused by insufficient training samples, such as inadequate identity diversity and restricted realism. Specifically: first, a finger vein identity generator models vascular topology under physiological and geometric constraints using stochastic L-systems, producing anatomically valid and identity-distinctive vascular patterns. Then, a cascaded region-aware GAN renders the topological maps into realistic near-infrared images. Finally, an intra-class diversity generator introduces geometric and optical perturbations to simulate realistic intra-class variations. Using FVeinSyn, we generated 500,000 images (10,000 vein identities, 50 samples per identity) and conducted extensive evaluations. Results show that FVeinSyn holds significant advantages in realism, identity diversity, vascular pattern consistency, and intra-class diversity. Models trained with FVeinSyn outperform real-data-only baselines a
cross eight public datasets, achieving an average accuracy improvement of 27.43\%. The code is available at: \url{https://github.com/EvanWang98/Synthetic-Finger-Vein-Generator}.
\end{abstract}

\begin{IEEEkeywords}
Synthetic finger vein generation, biometrics, lindenmayer systems, adversarial generation, finger vein recognition
\end{IEEEkeywords}

\section{Introduction}

\IEEEPARstart{O}{ver} the past two decades, biometric recognition technologies have achieved remarkable progress and have been widely deployed in scenarios such as security authentication, financial payment, and public safety~\cite{kim202550}. As the first-generation modalities, face~\cite{10.1145/3664647.3680635, 10399793} and fingerprint~\cite{dong2025bridging} recognition have benefited from large-scale datasets\cite{9763004, 9893541, sun2023zjut} and deep representation learning, reaching industrial-level maturity. However, they also expose significant security and privacy~\cite{10.1145/3664647.3680704, cai2025rehearsal} concerns: facial images can be easily captured and misused in public scenarios, while fingerprints are permanently exposed and thus vulnerable to duplication and spoofing. In contrast, finger vein recognition leverages subcutaneous vascular patterns, offering intrinsic resistance to forgery and stronger privacy protection~\cite{10620353}. Consequently, it has been recognized as a representative modality of the second-generation biometrics and has already seen industrial adoption (e.g., products from Hitachi, Fujitsu, and Tencent).

\begin{figure}[t]
	\centering
	\includegraphics[scale=1,width=0.48\textwidth]{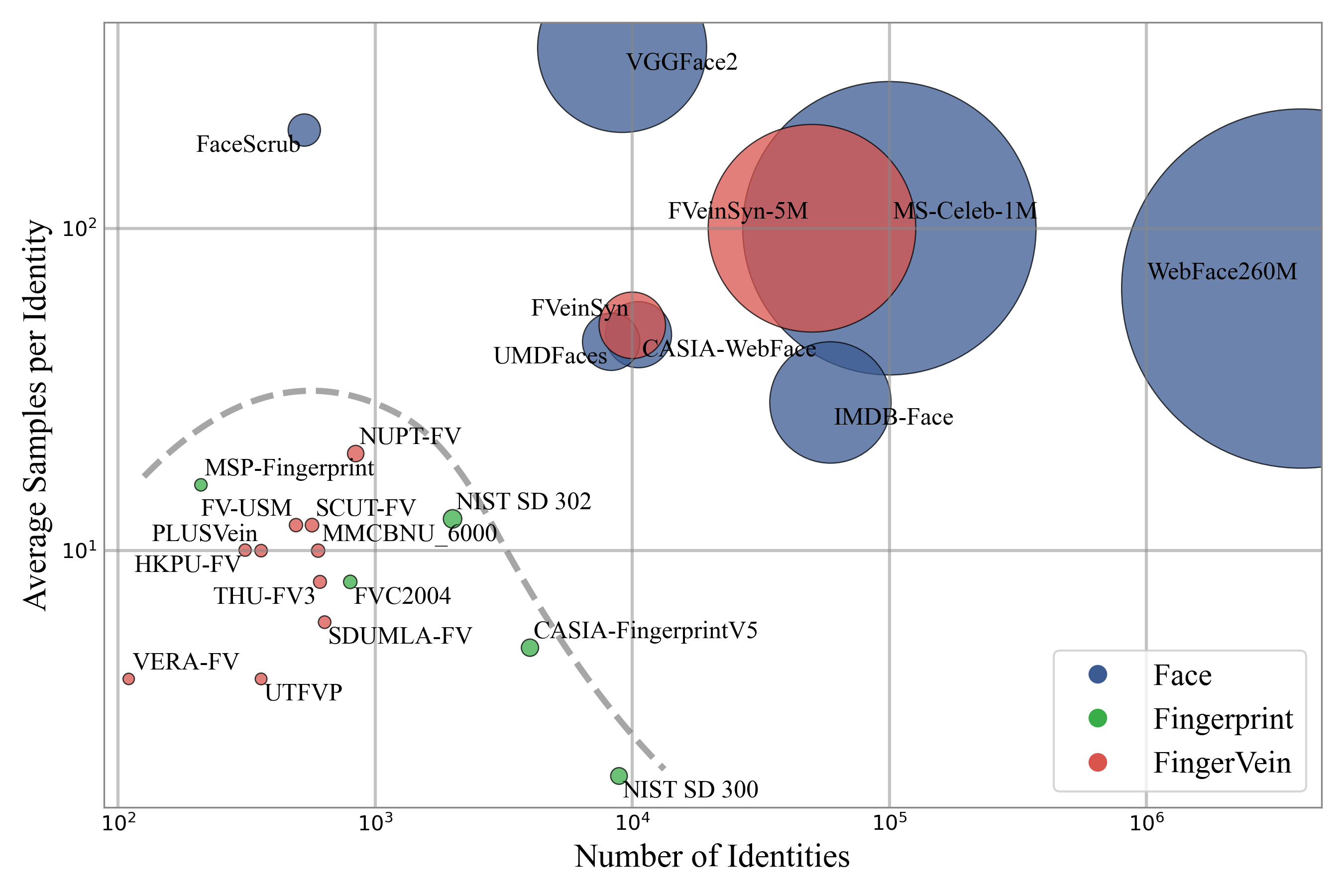}
	\caption{Comparison of publicly available datasets representing face~\cite{casia-webface, Guo2016MSCeleb1MAD, cao2018vggface2, bansal2017umdfaces, FaceScrub, wang2018devil, 9763004}, fingerprint~\cite{maio2004fvc2004, arakala2013msp, nist302, casiafingerprintv5, fiumara2018nist}, and finger vein~\cite{yin2011sdumla, Ton_UTFVPDB2013, lu2013available, 8698588, asaari2014fusion, kumar2011human, tang2019finger, 6467066, vanoni2014cross, ren2022dataset} biometric modalities. FVeinSyn-5M (50,000 identities/5,000,000 samples)~\cite{10.1145/3746027.3758253} and FVeinSyn (10,000 identities/500,000 samples) are synthetic finger vein datasets proposed in this work.}
	\label{fig1}
\end{figure}

Despite its advantages, the development of finger vein recognition is still constrained by the severe lack of large-scale public datasets~\cite{WANG2023119874}. Unlike face and fingerprint images, finger vein acquisition requires active participation and a more complex capture process, making it difficult to collect large-scale samples from the Internet~\cite{9763004}. As a result, current finger vein datasets fall far behind face and fingerprint datasets in terms of identity scale, per-identity sample size, and intra-class diversity (see fig.~\ref{fig1}). For instance, while face datasets now include over 4M unique identities and 260M images—enabling the success of deep representation learning—and fingerprint datasets reach nearly 10K identities, the largest available finger vein dataset still contains fewer than 1K identities. This disparity directly hinders research on critical issues such as million to billion scale retrieval, cross-device generalization, and robustness evaluation.

Publicly available finger vein datasets include SDUMLA-FV~\cite{yin2011sdumla}, UTFVP~\cite{Ton_UTFVPDB2013}, MMCBNU\_6000~\cite{lu2013available}, FV-USM~\cite{8698588}, HKPU-FV~\cite{asaari2014fusion}, PLUS-FV3~\cite{kumar2011human}, SCUT-FV3~\cite{tang2019finger}, THU-FV3~\cite{6467066}, VERA-FV~\cite{vanoni2014cross}, and NUTP-FV~\cite{ren2022dataset}. While they have provided valuable support for research, several limitations remain: (1) the number of unique identities is limited—fewer than 1K even in the largest dataset; (2) per-identity samples are scarce, typically ranging from 2 to 12 impressions per finger; (3) intra-class variations are minimal, as samples collected within the same session show almost no variation, with at most two sessions per dataset; and (4) auxiliary annotations such as joint cavity positions, ROI bounding box, vein pattern, and finger shape are unavailable, restricting the use of advanced deep learning techniques.

In various domains such as face and fingerprint recognition, synthetic datasets have been leveraged to improve model generalization under limited real data. Generative adversarial networks (GANs)~\cite{9893541} and diffusion models~\cite{10204758, jin2025diffpalm, grosz2024universal} have demonstrated strong image synthesis capabilities, effectively expanding training dataset. However, finger vein synthesis is fundamentally more challenging than generic image generation, because a high-quality synthesizer must simultaneously support large-scale identity creation, realistic intra-class variation simulation, and stable identity-label consistency. More importantly, unlike prior synthetic finger vein generation methods that either rely on explicit modeling of vascular geometry and optical priors or directly learn image-level mappings from limited real data, the proposed FVeinSyn adopts a structurally decoupled generation paradigm: vascular topology is generated explicitly in identity space, while near-infrared appearance is rendered separately in image space. This decoupling mechanism improves controllability and avoids forcing a single end-to-end generator to jointly learn complex vascular topology, finger geometry, and near-infrared imaging distributions from scarce real samples.

\begin{figure}[t]
	\centering
	\captionsetup[subfloat]{labelsep=space}
	
	\subfloat{
		\includegraphics[width=0.48\linewidth]{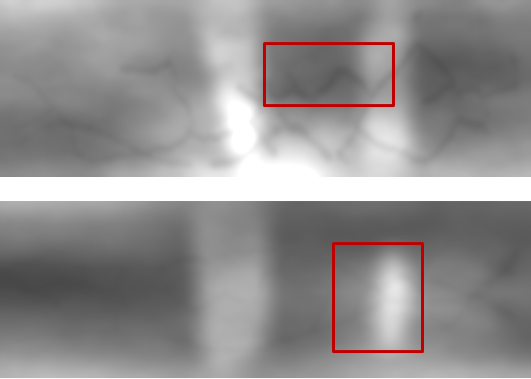}}
	\hfill
	\subfloat{
		\includegraphics[width=0.48\linewidth]{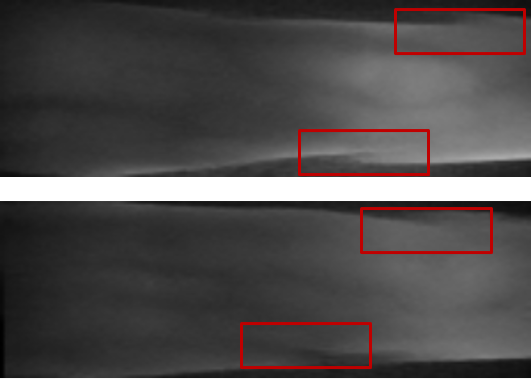}}
	
	\caption{Traditional modeling (left): vessels are unsmooth and rigid, with unrealistic infrared appearance. Learning-based (right): noticeable artifacts are observed.}
	\label{fig2}
\end{figure}

Traditional modeling of synthetic finger vein images primarily relies on explicit modeling methods based on geometric structures and optical priors~\cite{hillerstrom2014generating}. Although these approaches can reproduce the overall vascular morphology, they often fail to generate smooth and physiologically coherent local details, resulting in vascular patterns that appear rigid or unnatural. Moreover, empirical imaging models cannot fully capture the complex physical processes involved in near-infrared imaging, such as subsurface scattering and tissue absorption, leading to a gap in realism between synthetic and real images. Fig. \ref{fig2} shows limitations in existing synthetic finger vein generation methods.

From a deep generative modeling perspective, data scarcity poses unique challenges to finger vein generation. The limited real samples cannot adequately cover the broad variability of vascular topologies, nor are they sufficient for a single end-to-end generator to simultaneously learn both the identity distribution (topology) and the appearance distribution (near-infrared imaging style)~\cite{ou2022gan}. Moreover, generating sufficiently diverse vein patterns alone is already a challenging task; jointly modeling vein patterns, imaging characteristics, and finger geometry further increases the complexity. In particular, under limited data conditions, it remains extremely difficult to stably learn finger vein image samples with synthetically generated anatomically plausible finger geometries from scratch.

To address these challenges, we propose FVeinSyn, a large-scale and controllable synthetic finger vein image generation framework. FVeinSyn explicitly decouples identity generation (topology) and near-infrared appearance rendering, and further introduces controllable structural and degradation priors to model realistic complete finger vein images with rich inter-class and intra-class variations.

Specifically, we propose a vascular identity generator, which defines a stochastic vascular growth grammar constrained by anatomical and topological priors and implements it within a symbolic L-system framework. The generator models vascular formation as a recursive evolutionary process governed by probabilistic production rules, inspired by the biological mechanism of vascular bifurcation. By incorporating physiological constraints such as branching angles, diameter decay rates, and curvature smoothness during generation, the model produces vascular patterns that are both topologically continuous and physiologically plausible, overcoming the local rigidity and morphological discontinuity issues of traditional explicit modeling methods. Furthermore, it enables the synthesis of anatomically realistic binary vascular patterns in topological space with controllable and identity-distinctive priors.

We further propose a region-aware cascaded adversarial network as the synthetic image renderer. It takes binary vascular topology as input, allowing the generator to bypass the need to learn complex vascular topology distributions in image space and instead focus on modeling the near-infrared appearance style under the supervision of a limited number of real vein samples. Furthermore, we assign each vascular identity a unique set of anatomical structural priors—including finger shape, joint cavity regions, and regions of interest—to enhance identity distinctiveness. These structural priors can be obtained in a controllable and low-cost manner, making their integration both practical and scalable for large-scale synthesis. Guided by these priors, the renderer maps binary vein patterns to realistic near-infrared appearances. In addition, a cascaded regional supervision mechanism enforces multi-region consistency, thereby improving both the global realism and local fidelity of the synthesized images.

Finally, unlike existing synthetic finger vein generation methods that typically produce only localized vein-region images with subtle variations~\cite{ou2022gan,hillerstrom2014generating,shang2025pvtree,Shen_2023_ICCV,zhao2022bezierpalm}, FVeinSyn generates complete finger vein images and explicitly models intra-class diversity. To this end, we introduce an intra-class diversity generator, which incorporates geometric transformations (e.g., translation, rotation, and scaling), photometric changes (e.g., over/under exposure), and imaging degradations (e.g., motion blur, optical blur, and scattering blur) to more realistically simulate the complexity and diversity of real finger vein data.

The followings are the contributions of the paper.

\begin{itemize}
	\item We propose FVeinSyn, the first large-scale, anatomically grounded, and controllable synthetic data generation framework for finger vein. It enables identity-consistent and realistic vein synthesis at scale without relying on large real samples.
	
    \item We propose a vascular identity generator that defines a novel stochastic vascular growth grammar constrained by finger physiological priors and implements it within a symbolic L-System formalism to generate anatomically plausible and identity-distinctive vascular topologies.
	
	\item We propose a region-aware cascaded renderer that maps binary vascular topologies into realistic near-infrared images through multi-region adversarial generation.
	
    \item We propose an intra-class diversity generator that simulates realistic intra-class variability through controllable geometric, photometric, and imaging transformations.
	
    \item Extensive experiments demonstrate that FVeinSyn-generated datasets improve downstream recognition performance, outperforming existing state-of-the-art finger vein generation methods in identity scale and diversity.
\end{itemize}

Our earlier conference work~\cite{10.1145/3746027.3758253} released FingerVeinSyn-5M, the largest synthetic finger vein dataset to date, which was generated using FVeinSyn. That work primarily focused on the construction of the dataset and its preliminary evaluation. Compared with the conference version, this study provides a detailed description of the complete FVeinSyn synthetic finger vein image generation framework, elaborates on the design and implementation of its key components, including the Vascular Identity Generator, Region-Aware Cascaded Renderer, and Intra-Class Diversity Generator, and presents a significantly expanded evaluation of the generation framework.

\section{Related Work}

\subsection{Finger Vein Recognition}
Finger vein recognition initially relied on handcrafted feature extraction methods, which use manually designed pattern descriptors and matching algorithms to classify vein identifies. Representative methods include repeated line tracking (RLT)~\cite{RLT}, maximum curvature (MC)~\cite{MC}, Gabor filtering~\cite{kumar2011human}, isotropic undecimated wavelet transform (IUWT)~\cite{IUWT}, and wide line detector (WLD)~\cite{WLD}. Although these traditional methods can achieve basic recognition performance in controlled environments, their effectiveness is limited by insufficient feature representation and poor adaptability to environmental changes. They are sensitive to variations in lighting, finger misalignment, and imaging noise~\cite{wang2025multi}. The choice of matching strategy also affects performance.

With the development of deep learning, end-to-end training methods based on convolutional neural network have significantly improved recognition accuracy and system robustness, driving a paradigm shift in the field~\cite{radzi2016finger, 8395431}. Current deep learning approaches can be categorized into several aspects: network architecture design~\cite{10023509, 8979362}, loss functions~\cite{9363648}, prior knowledge~\cite{WANG2023119874}, preprocessing~\cite{10620353}, domain adaptation and robustness.

Firstly, for the architecture, researchers have gradually moved from shallow convolutional network to deeper and more expressive backbones, such as residual network~\cite{WANG2023119874}, lightweight~\cite{shen2021finger}, and multi-scale branches~\cite{qin2021multi, pan2020multi}. To address the inherently low contrast and fine pattern of finger veins, network designs often incorporate feature pyramids, full view~\cite{huang2024mirror} and attention mechanisms~\cite{qin2024attention, qin2022local} to enhance the response and localization of small vascular patterns. At the same time, the demand for lightweight and efficient inference has led to the development of compact network and pruning/quantization strategies that can be deployed on resource-limited devices~\cite{shen2021finger}.

Secondly, in terms of loss functions and training paradigms, metric learning approaches (contrastive loss, triplet loss, center loss)~\cite{9408606, wang2025colorvein}  and angular/margin-based losses (ArcFace, CosFace)~\cite{9363648} have become mainstream methods for improving discriminative feature representation, significantly increasing inter-class separation and intra-class compactness. Recently, self-supervised~\cite{ou2024gscl} and contrastive learning~\cite{ma2023focal, ou2022gan} methods have been applied to finger vein recognition to learn robust features. Meta learning~\cite{tang2019finger} have also been explored for sample limited recognition.

Despite significant improvements on several benchmarks, existing deep learning methods still face performance limitations~\cite{9973284}. First, the scale and diversity of available data are insufficient; the number of identities, the number of samples per identity, and the number of acquisition sessions in public datasets are far from meeting the requirements for learning broad distributions in deep models. Second, intra-class variations are complex, with posture, light scattering, partial occlusion, and differences in imaging devices all contributing to high-dimensional variations that simple augmentation and end-to-end training cannot fully capture~\cite{ren2025iis}. These challenges directly limit the model’s generalization ability in cross-domain, few-shot, and large-scale retrieval scenarios~\cite{10530126}.

\subsection{Synthetic Finger Vein Generation}
In recent years, the generation of synthetic biometric data has become an important direction for enhancing the generalization and scalability of recognition models. In the field of finger vein recognition, existing studies can be broadly divided into two categories: intra-class generation~\cite{10530126, WANG2023119874} and inter-class generation~\cite{ou2022gan, hillerstrom2014generating, 10.1145/3746027.3758253}. The former aims to produce diverse supplementary samples for known identities to improve intra-class robustness, while the latter focuses on constructing new identities along with their corresponding multi-modal samples to expand the identity space and support model pretraining. Currently, most research still concentrates on intra-class generation, with only a few studies attempting to address the inter-class generation problem.

Hillerstr{\"o}m~\emph{et al.} ~\cite{hillerstrom2014generating} first explored an analytical modeling-based approach for synthesizing finger vein images and released a synthetic dataset containing about 5,000 identities, each with 10 samples. This pioneering study laid the foundation for subsequent research, but it still showed clear limitations in visual realism, geometric continuity, and intra-class diversity, as its variations were mainly introduced through limited parameter perturbations. Later, Ou \emph{et al.}~\cite{ou2022gan} proposed an image generation framework based on ROI block shuffling and generative adversarial network, which achieved more natural pattern appearance through region remapping and adversarial rendering. However, due to insufficient training samples, most of the generated sample contain artifacts.

In contrast, related biometric modalities such as palm vein and palmprint synthesis have achieved more systematic progress, with approaches based on morphological modeling, cross-modal style transfer, and conditional generation~\cite{zhao2022bezierpalm, jin2025diffpalm, Shen_2023_ICCV, jin2024pce, shang2025pvtree}. Nevertheless, the synthesis of finger vein images remains more challenging. First, it requires accurate reconstruction of fine anatomical structures of real fingers (such as joint cavities and finger shape). Second, the imaging process is affected by complex optical effects, including near-infrared light transmission, subsurface scattering, and local occlusion—factors that make generation far more difficult than for palm or dorsal hand regions. Existing research mostly focuses on local ROI-level synthesis, whereas high-fidelity, full-finger generation has greater potential for identity modeling, structural prior learning, and multi-task recognition~\cite{ou2022gan, hillerstrom2014generating}.

\subsection{Biometrics with Synthetic Dataset}
Compared with real biometric data collected from individuals, synthetic data offers significant advantages in terms of ethical compliance and distributional balance~\cite{9209125}. In recent years, large-scale real datasets have raised growing concerns regarding privacy protection, lack of informed consent, and racial bias~\cite{10204758}. In contrast, synthetic data can be expanded at scale without violating personal privacy, effectively mitigating such risks. Moreover, the generation process of synthetic samples can be strictly controlled, allowing researchers to precisely balance factors such as identity distribution, gender ratio, and imaging conditions, thereby reducing the impact of class imbalance on model learning~\cite{yao2025synthetic}.

Nevertheless, recognition systems trained solely on synthetic data still face performance limitations. The main reason lies in the texture fidelity and domain distribution gap of generated samples, which restrict their generalization capability to real-world test sets. Therefore, in most biometric recognition tasks, synthetic data is typically used as a supplement to real data. In face recognition~\cite{10204758, flotho2025t, 9737477}, fingerprint recognition~\cite{dong2025bridging, dong2023synthesis, 9893541, grosz2024universal}, palmprint recognition~\cite{zhao2022bezierpalm, jin2025diffpalm, Shen_2023_ICCV, jin2024pce}, and anti-spoofing tasks~\cite{fan2025divtrackee, grosz2022spoofgan, sun2020face}, numerous studies have demonstrated that joint training with real and synthetic samples can effectively improve both recognition accuracy and robustness.

This trend indicates that the value of synthetic data is shifting from simple data augmentation toward task-consistent enhancement—that is, improving the generalization boundary of feature representation through complementary distributions with real samples. Consequently, high-fidelity, controllable, and task-aligned synthetic data is expected to become a key foundation for advancing the next generation of biometric recognition systems.
\begin{figure}[!t]
	\centering
	\includegraphics[width=0.9\linewidth]{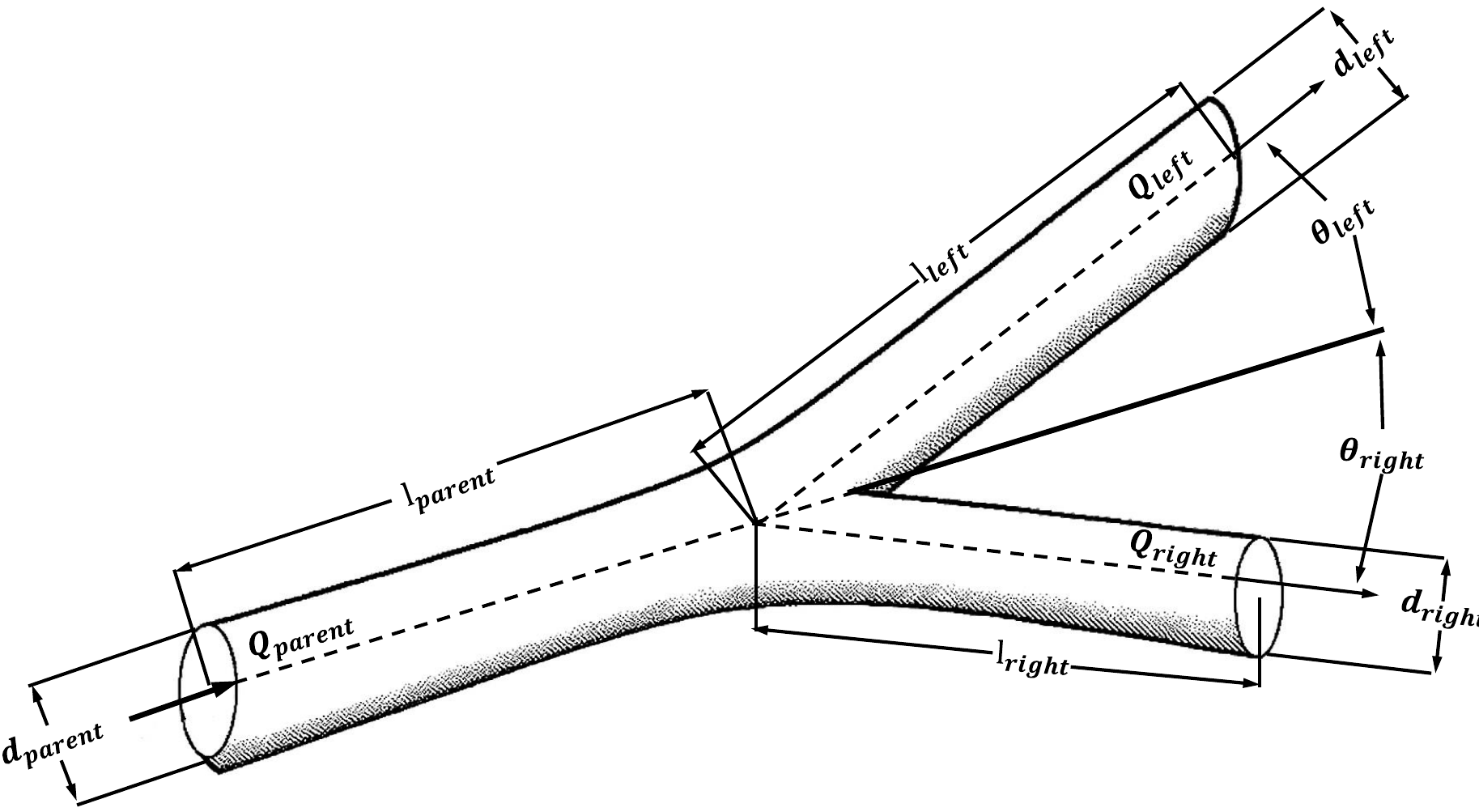}
	\caption{\textbf{Bifurcation parameters:} diameters ($d_{\mathrm{parent}}$, $d_{\mathrm{left}}$, $d_{\mathrm{right}}$), lengths ($l_{\mathrm{parent}}$, $l_{\mathrm{left}}$, $l_{\mathrm{right}}$), volumetric flows ($Q_{\mathrm{parent}}, Q_{\mathrm{left}}, Q_{\mathrm{right}}$) and bifurcation angles ($\theta_{\mathrm{left}}$, $\theta_{\mathrm{right}}$).
	}
	\label{fig2}
\end{figure}
\section{Proposed Approach}
\begin{figure*}[t]
	\centering
	\includegraphics[scale=1,width=0.98\textwidth]{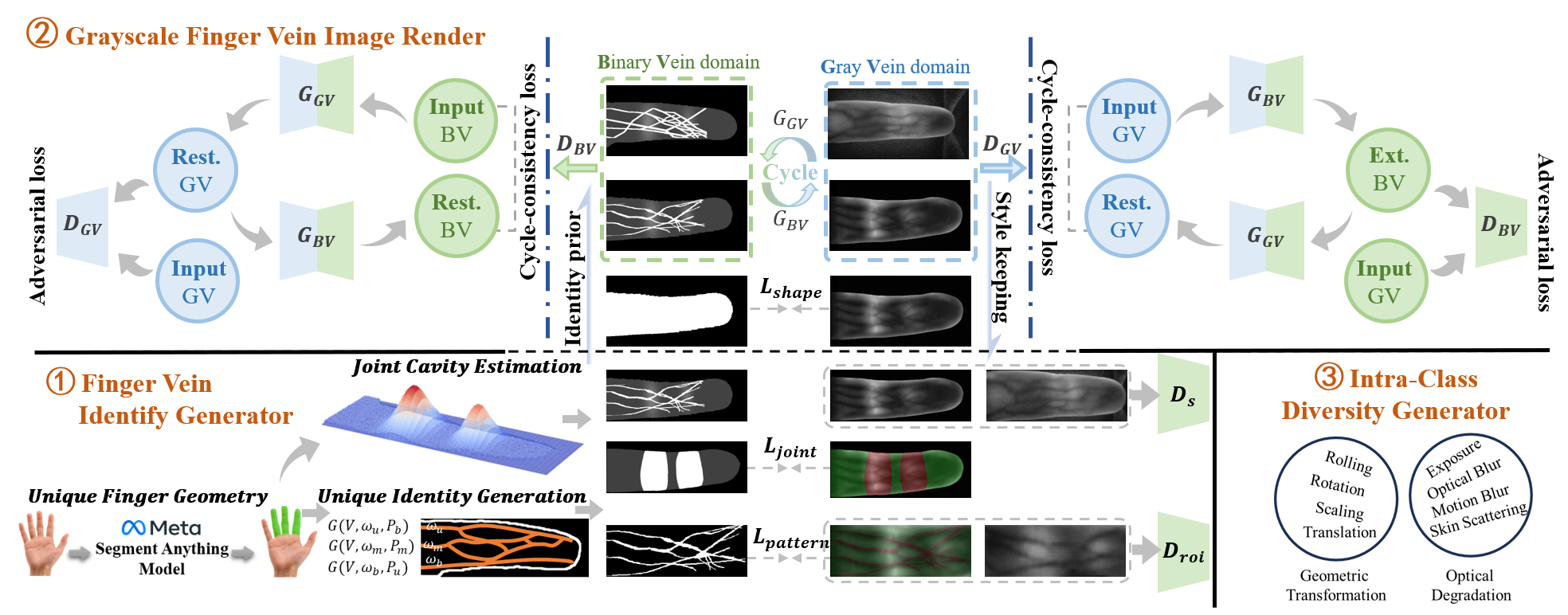}
	\caption{Overview of the FVeinSyn framework for large-scale synthetic finger vein image generation. 1) Finger Vein Identity Generator, where unique vascular topologies are created. 2) Grayscale Finger Vein Image Render, performed by a Cascaded Region-Aware GAN (CascadedRA-GAN) to produce realistic images. 3) Intra-Class Diversity Generation, which applies geometric and optical perturbations to enhance variability within each identity.}
	\label{fig3}
\end{figure*}

Our proposed framework synthesizes realistic finger vein images through a three stage process. Fig.~\ref{fig3} illustrates the overview of the FVeinSyn framework. First, a binary vascular topology map $I_P$ is generated, which recursively simulates the branching of finger veins under physiological and geometric constraints. Next, the generated topology map $I_P$ is fed into a cascaded region-aware generative adversarial network (CascadedRA-GAN) along with a set of structural priors $\mathcal{S}$, including finger shape masks and joint-cavity annotations. This module renders $I_P$ into a realistic near-infrared image $I_{syn}$. Finally, to further enhance intra-class diversity, the intra-class diversity generator (ICDG) applies a sequence of geometric and optical perturbations in a probabilistic manner, simulating pose variations, skin scattering, exposure changes, and imaging blur. Thus, by FVeinSyn framework can synthesize a wide range of realistic finger vein images for diverse identities and acquisition conditions. Each of these components is described in detail in the subsections below.

\subsection{Preliminary}
\noindent
\textit{\textbf{Vascular Network}}. 
The topology and geometry of vascular network can be characterized by a set of fundamental physical--geometric constraints, primarily involving flow conservation and a power-law relationship of branch radii~\cite{zamir1976optimality, sweeney2024unsupervised}. At a bifurcation, let the volumetric flows of the parent vessel and the left and right child vessels be $Q_{\mathrm{parent}}, Q_{\mathrm{left}}, Q_{\mathrm{right}}$, respectively, and their radii be $d_{\mathrm{parent}}, d_{\mathrm{left}}, d_{\mathrm{right}}$. The branching asymmetry ratio is defined as
$\alpha= d_{\mathrm{right}}/d_{\mathrm{left}},$
and the left and right child branches form inclination angles $\theta_{\mathrm{left}}, \theta_{\mathrm{right}}$ relative to the parent segment, as illustrated in Fig.~\ref{fig2}.

\noindent
\textit{Flow Conservation} requires that the incoming flow equals the total outgoing flow:
\begin{equation}
Q_{\mathrm{parent}} = Q_{\mathrm{left}} + Q_{\mathrm{right}},
\end{equation}
meanwhile, vessel radii satisfy a \textit{Power-Law Constraint}, commonly referred to as the Murray-type law:
\begin{equation}
d_{\mathrm{parent}}^{\gamma} = d_{\mathrm{left}}^{\gamma} + d_{\mathrm{right}}^{\gamma}
\end{equation}
where the branching exponent $\gamma$ is typically in the range $2.55$--$3.0$, with the classical case $\gamma \approx 3$ corresponding to Murray’s law~\cite{murray1926physiological}. The asymmetry ratio $\alpha$ quantifies the imbalance between the two child branches.

During vascular tree generation or modeling, each vessel segment grows until it reaches a threshold length $l$, at which point it bifurcates into two child segments oriented by angles $\theta_{\mathrm{left}}$ and $\theta_{\mathrm{right}}$. These constraints jointly ensure the consistency of blood flow and geometric scaling in vascular network.

\noindent
\textit{\textbf{Lindenmayer Systems}}
(L-systems) are formal grammar systems that provide a mathematical framework for modeling the growth processes of organisms, particularly plants~\cite{lindenmayer1968mathematical}. A context-free L-System (0L-System) applies productions without considering the context of symbols within a string. Formally, it is defined as a triple $L = (V, P, \omega)$, where $V$ is a finite set of symbols (alphabet) that may include terminal and non-terminal symbols; $P$ is a finite set of production rules, each of the form $(A, \alpha) \in P$, denoted as $A \to \alpha$, where $A$ is the predecessor and $\alpha$ is the successor string; and $\omega \in V^*$ is the axiom, serving as the initial string for the derivation process. Geometrically, 0L-Systems are typically interpreted by mapping symbols to drawing instructions, thereby generating complex structures such as fractals or branching patterns. For instance, a symbol may represent moving forward in a given direction or rotating by a specified angle in the drawing plane.

When each symbol $A$ has exactly one associated production rule, the 0L-System is referred to as deterministic (D0L-System). Conversely, if a function assigns a probability to each production, the resulting system is a context-free stochastic L-System (S0L-System), which is the focus of this work~\cite{prusinkiewicz2013lindenmayer}. In S0L-Systems, probabilities are introduced to production rules, rendering the derivation process non-deterministic. Formally, the system is represented as a quadruple $L = (V, P, p, \omega)$. For any predecessor symbol $A$, the sum of probabilities of all corresponding productions equals one,
\begin{equation}
\sum_{A \to \alpha \in P} p(A \to \alpha) = 1.
\end{equation}

The probabilistic mechanism of S0L-Systems introduces randomness into the derivation process. In addition, production rules can be parameterized to simulate natural variations in growth patterns, such as adjusting branching angles and segment lengths, enabling the generation of realistic synthetic vascular network.

\noindent
\textit{\textbf{Adversarial Image-to-Image Style Transfer}}.
%\textbf{Notation and Objective.} 
Let $X$ and $Y$ denote two image domains with real data distributions $p_X$ and $p_Y$, respectively. Adversarial image-to-image style transfer aims to learn a pair of mappings
$G: X \rightarrow Y, \quad F: Y \rightarrow X$,
along with discriminators $D_Y$ and $D_X$ that distinguish whether samples belong to the target domain. The main goal of this class of methods is to perform cross-domain image translation while preserving the semantic and geometric content of the input images, often without requiring paired training samples~\cite{zhu2017unpaired}.

%\textbf{Adversarial Loss.} 
To ensure that the generated images $G(x)$ follow the target distribution $p_Y$, an adversarial loss is defined as
\begin{equation}
\mathcal{L}_{\text{GAN}}^{X \to Y} = \mathbb{E}_{y}[(D_Y(y)-1)^2] + \mathbb{E}_{x}[D_Y(G(x))^2].
\end{equation}

%\textbf{Cycle-Consistency Loss.} 
Since relying solely on adversarial loss may result in degenerate mappings or content loss, a cycle-consistency constraint is introduced:
\begin{equation}
	\small
\mathcal{L}_{\text{cyc}} = \mathbb{E}_{x \sim p_X}[\|F(G(x))-x\|_1] + \mathbb{E}_{y \sim p_Y}[\|G(F(y))-y\|_1].
\end{equation}
This encourages $F \circ G \approx \text{id}_X$ and $G \circ F \approx \text{id}_Y$, thereby preserving input content and enabling invertible mappings.

\subsection{Finger Vein Identity Generator}
Finger vein patterns exhibit inherently individual-specific vascular structures, whose branching topology and spatial distribution remain highly stable across multiple acquisitions, yet differ significantly between individuals. Features such as bifurcation geometry, vessel diameter ratios, and directional organization form the structural foundation of biometric recognition. Therefore, the synthesis of vein identities must balance anatomical plausibility and topological diversity, ensuring that each identity corresponds to a physiologically valid yet unique vascular pattern.  

To this end, we model vascular formation as a novel probabilistic growth grammar whose production rules correspond to biologically meaningful branching events, where rule probabilities are parameterized by anatomical and geometric priors of the finger, implementing it within a symbolic L-System framework to generate anatomically plausible and identity-distinctive vascular topologies. Specifically, the generation starts from an initial trunk at the finger base, and child branches are recursively created according to stochastic production rules. The grammatical derivation produces a symbolic string, in which each symbol is mapped to a local drawing primitive---such as ``move forward by length $l$,'' ``rotate by angle $\theta$,'' or ``generate branch.'' Sequential execution of these primitives yields a set of 2D line-segment coordinates, known as \textit{turtle trajectories}. Bézier interpolation is then applied to these trajectories to form smooth and continuous vascular curves, which are subsequently rasterized into a binary image $I_P$, where the vessel thickness is determined by the average diameter of each segment.  

We describe the generation process from the perspective of a single vascular trunk and its recursive bifurcations. Beginning with an initial segment of length $l_0$ and diameter $d_0$, the model grows recursively following parameterized stochastic production rules. For a parent vessel $F(l)$ of length $l$. At each recursion step, a branching production rule is applied with probability $p \in (0,1)$, where $p$ is a fixed model parameter controlling local branching frequency. A typical production rule can be expressed as:
\begin{equation}
% F(l) \rightarrow p\,F(l\cdot\eta_l)[+\theta]\,F(l\cdot\eta_l)[-\theta],
F(l) \rightarrow p\, [+\theta\,F(l\cdot\eta_l)]\,[-\theta\,F(l\cdot\eta_l)],
\end{equation}
where $\eta_l \in (0,1)$ denotes the inter-level length attenuation factor and $\theta$ represents the branching angle. The bracket operators $[\,\cdot\,]$ indicate push and pop operations on the growth state stack, which stores the current position and orientation. With probability $1-p$, the vessel continues elongation without branching. Here, $\eta_l \in (0,1)$ denotes the inter-level length attenuation factor. The vessel diameter is attenuated by a scaling factor $\eta_d$ at each growth step, which is chosen to approximately satisfy the Murray-type law~\cite{murray1926physiological}.

Through this process, the generator produces vascular network with clear hierarchical organization and diverse topological structures. Each synthetic identity typically contains three to four primary trunks (distributed vertically along the finger axis) and their secondary branches, exhibiting physiologically consistent asymmetry and density distributions. Finally, the generated vascular structure is rasterized into a binary \textit{topology map} $I_P$, serving as the foundational representation for finger vein identity generation and providing anatomically consistent, controllable priors for the subsequent image rendering stage.

\subsection{Finger Vein Image Renderer: CascadedRA-GAN}
The synthesis of realistic finger vein images can be formally described as a conditional generative modeling problem, where the objective is to map a binary vein identity pattern \(I_p\) into a realistic near-infrared image while preserving both anatomical plausibility and inter-subject variability. To this end, we propose a cascaded region-aware generative adversarial network (CascadedRA-GAN), whose central idea is to decompose the synthesis process into two coupled sub-tasks: 
(i) mapping binary vein patterns to realistic appearance under structural priors; and 
(ii) enforcing multi-region consistency through cascaded regional supervision, thereby enhancing global realism and local fidelity simultaneously. 

Specifically, given a binary vein identity pattern \(I_p\), the generator further takes as input a set of structural constraints:
\begin{equation}
\mathcal{S} = \{ I_s, D_j, A \},
\end{equation}
where \(I_s\) denotes the finger shape mask, \(D_j\) represents the brightness distribution of joint cavities, and \(A\) comprises auxiliary annotations including joint cavity coordinates, region of interest, and local deformation descriptors. The rendering task can thus be formulated as learning the mapping:
\begin{equation}
G:(I_p, \mathcal{S}) \mapsto I_{\text{syn}},
\end{equation}
where \(I_{\text{syn}}\) denotes the synthesized finger vein image. 

In contrast to approaches relying on approximate or hand-crafted priors, our design explicitly incorporates real anatomical structures as generative constraints. Concretely, we leverage a large-scale gesture database~\cite{Kapitanov_2024_WACV} with the \emph{Segment Anything Model}~\cite{kirillov2023segany} to achieve precise segmentation of finger regions, thereby producing explicit masks for ROI, joint cavities, and overall finger. This ensures that each synthesized vein identity is embedded within a unique and anatomically consistent finger structure, effectively preventing mode collapse and enhancing both inter-subject diversity and realism.

To balance global consistency with local structural fidelity, we design a cascaded region-aware loss for the proposed CascadedRA-GAN. The overall objective is defined as:
\begin{equation}
	\mathcal{L} = \mathcal{L}_{\text{CycleGAN}} + \mathcal{L}_{\text{Region-Aware}},
\end{equation}
where $\mathcal{L}_{\text{CycleGAN}}$ includes standard identity, cycle-consistency, and adversarial losses to ensure distributional alignment between synthesized and real samples at global scale.

The region-aware loss $\mathcal{L}_{\text{Region-Aware}}$ is defined as
\begin{equation}
	\small
	\mathcal{L}_{\mathit{RA}} = \mathcal{L}_{\mathit{cycle}}^s + \mathcal{L}_{\mathit{cycle}}^{\mathit{roi}} + \mathcal{L}_{\mathit{adv}}^s + \mathcal{L}_{\mathit{adv}}^{\mathit{roi}} + \lambda_1 \mathcal{L}_{\mathit{s}} + \lambda_2 \mathcal{L}_j + \lambda_3 \mathcal{L}_p,
\end{equation}
where $\mathcal{L}_{\text{cycle}}^{s}$ and $\mathcal{L}_{\text{cycle}}^{roi}$ denote cycle-consistency constraints imposed on the finger-shape region and the ROI, respectively, while $\mathcal{L}_{\text{adv}}^{s}$ and $\mathcal{L}_{\text{adv}}^{roi}$ represent the corresponding regional adversarial losses. These local discriminators operate independently of the global discriminator, thereby capturing fine-grained style variations at multiple scales.

In addition, the shape-consistency loss $\mathcal{L}_{\text{s}}$ is based on the binary finger-shape mask $I_{s}$, where Hausdorff distance (Haber loss) is employed to constrain the contour alignment. The joint-cavity loss $\mathcal{L}_{j}$ and the vein-pattern loss $\mathcal{L}_{p}$ are inspired by physiological priors of near-infrared imaging. Specifically, joint-cavity regions generally exhibit higher average brightness, whereas vein regions appear darker. To model this explicitly, we introduce a contrastive loss:
\begin{equation}
	\small
	\mathcal{L}_{j/p} = \frac{1}{N} \sum_{i=1}^{N} \max \left( 0, \ \text{Margin} - \big(\mu_A(x_i) - \mu_B(x_i) \big) \right),
\end{equation}
where $x_i$ denotes the $i$-th generated image, and $\mu_{A}(\cdot)$ / $\mu_{B}(\cdot)$ indicate average pixel intensity within regions $A$ and $B$, respectively. In $\mathcal{L}_j$, region $A$ denotes the finger joint cavity region obtained from $D_j$, while region $B$ denotes the finger mask region obtained from $I_s$. 
In $\mathcal{L}_p$, region $B$ corresponds to the vein pattern region obtained from $I_p$, and region $A$ represents the non-vein region, as illustrated in Fig.~\ref{fig3}. For joint cavities and vein regions, we set $\text{Margin}_j = \text{Margin}_p = 0.2$, thereby enforcing explicit discriminability in brightness distributions across tissue types.

This multi-level loss design ensures that the generator not only learns global distributional consistency but also captures fine-grained structural, brightness, and pattern constraints, thereby enhancing both realism and discriminability in cross-identity synthesis.

\begin{table*}[t]
	\centering
	\renewcommand{\arraystretch}{1.0}
	\caption{Parameterization for ICDG operators. $\mathcal{U}(a,b)$ denotes the uniform distribution over the interval $[a,b]$. $W$, $H$ are width and height, $\tau$ is the translation factor.}
	\label{tb1}
	\begin{tabular}{@{}lllcl@{}}
		\toprule
		Operator & Params & Range / distribution & Default & Notes \\ \midrule
		\multirow{2}{*}{Translation} 
		& $t_x$ & $U(-\tau_w W,\tau_w W)$ & $\tau_w=0.05$ & horizontal displacement \\ 
		& $t_y$ & $U(-\tau_h H,\tau_h H)$ & $\tau_h=0.05$ & vertical displacement \\ \midrule
		Rotation & $\phi$ & $U(-\phi_{\max},\phi_{\max})$ & $\phi_{\max}=15^\circ$ & rotation angle \\ \midrule
		Rolling & $\psi$ & $U(-\psi_{\max},\psi_{\max})$ & $\psi_{\max}=10^\circ$ & rolling angle \\ \midrule
		Scaling & $s$ & $U(1\pm s_{max})$ & $s_{m}=0.1$ & scale factor \\ \midrule
		Exposure & $v_0$ & $U(-v_{max},v_{max})$ & $v_{max}=2$ & exposure intensity \\ \midrule
		\multirow{2}{*}{Skin scatter} 
		& $\beta$ & $U(\beta_{min},\beta_{max})$ & $\beta_{min}=0.05, \beta_{max}=0.10$ & scattering coefficient \\ 
		& $\alpha$ & $U(\alpha_{min},\alpha_{max})$ & $\alpha_{min}=0.02, \alpha_{max}=0.06$ & decay rate \\ \midrule
		\multirow{2}{*}{Motion blur} 
		& $L$ & $Choice\{L_i\}$ & $\{L_i\}=\{1,2,\dots,15\}$ & kernel length \\ 
		& $\theta$ & $U(0,2\pi)$ & $\theta_{max}=2\pi$ & blur direction \\ \midrule
		\multirow{2}{*}{Optical blur} 
		& $K$ & $Choice\{K_i\}$ & ${K_i}=\{3,5,7,9,11,15\}$ & kernel size \\ 
		& $\sigma$ & $U(0,\sigma_{max})$ & $\sigma_{max}=10$ & standard deviation \\ \midrule
		Trigger ($\mathcal{G}$ / $\mathcal{O}$) & $\omega_k$ & weights $\omega_k$ & 
		\begin{tabular}[c]{@{}l@{}}
			$\omega_{\text{trans}}=0.8$, $\omega_{\text{rot}}=0.8$, $\omega_{\text{roll}}=0.5$, $\omega_{\text{scale}}=0.5$,\\
			$\omega_{\text{expo}}=0.3$, $\omega_{\text{scatter}}=0.8$, $\omega_{\text{motion}}=0.5$, $\omega_{\text{opt}}=0.5$
		\end{tabular} 
		& trigger probability \\
		\bottomrule
	\end{tabular}
\end{table*}

\subsection{Intra-Class Diversity Generator}
To enhance the intra-class diversity of finger vein image samples, we design a two-stage intra-class diversity generator (ICDG). Given an identity topology map $I_p$ and a structural prior $\mathcal{S}$, ICDG first applies geometric perturbations in the binary topology domain to simulate variations in pose and viewpoint. Subsequently, optical degradations are introduced in the grayscale rendering domain to mimic realistic near-infrared acquisition conditions, such as exposure, skin scattering, and imaging blur. Table~\ref{tb1} summarizes the parameter settings for each ICDG operator. 

We define a composite transformation associated with a transformation family $P$, which can be either the geometric family $\mathcal{G}$ or the optical family $\mathcal{O}$. 
The composite mapping is constructed as an ordered function composition of a subset of elementary transformations and is defined as

\begin{equation}
\begin{aligned}
T_P(\cdot;\boldsymbol{\theta}_P)
&=
T_{k_n}(\cdot;\theta_{k_n})
\circ
T_{k_{n-1}}(\cdot;\theta_{k_{n-1}}) \\
&\quad \circ \cdots \circ
T_{k_1}(\cdot;\theta_{k_1}),
\quad k_i \in P_{\mathrm{act}}.
\end{aligned}
\end{equation}

where $T_k(\cdot;\theta_k)$ denotes the $k$-th elementary transformation parameterized by $\theta_k$. 
$P_{\mathrm{act}}=\{k_1,\ldots,k_n\}\subseteq P$ represents the activated subset of transformations selected according to the triggering strategy, and 
$\boldsymbol{\theta}_P=\{\theta_{k_i}\}_{i=1}^n$ denotes the corresponding set of transformation parameters. So, the ICDG pipeline can be expressed as
\begin{equation}
\tilde{I}_p = T_{\mathcal{G}}(I_p; \theta_{\mathcal{G}}), \,\,
I_{syn} = G(\tilde{I}_p, \mathcal{S}), \,\,
\hat{I} = T_{\mathcal{O}}(I_{syn}; \theta_{\mathcal{O}}),
\end{equation}
where $\mathcal{G}=\{ {translation},  {rotation},  {rollnig},  {scaling}\}$ and $\mathcal{O}=\{ {exposure},  {skin\, scatter},{motion\, blur},  {optical\, blur}\}$. Based on the above formulation, we further define the parameterized sub-transformations $T_k(\cdot;\theta_k)$ for geometric perturbations and optical degradations as follows.

\noindent
\subsubsection{\textbf{\textit{Geometric Perturbations (Binary/Topology Stage)}}}

\noindent\textbf{Translation:} $T_{\text{trans}}(I;(t_x,t_y))$ will translate the finger within the plane, $t_x \sim U\left(-\tau_w W, \tau_w W\right)$, $t_y \sim U\left(-\tau_h H, \tau_h H\right)$.

\noindent\textbf{Rotation:} $T_{\text{rot}}(I;\phi)$ rotates the finger around its center within the plane, the angle sampling set to $\phi \sim U(-\phi_{\max}, \phi_{\max})$.

\noindent\textbf{Rolling:} $T_{\text{roll}}(I;\psi)$ simulates rolling effect of the finger by rotating around its longitudinal axis. 
Based on two assumptions: the cross-section of the finger is approximately elliptical, and the imaged veins are close to the finger's surfac, we employ an elliptical normalization scheme. 
Let $(u,v)$ denote the normalized elliptical coordinates:
\begin{equation}
u = \frac{x-x_c}{a}, \quad v = \frac{y-y_c}{b}, 
\end{equation}
where $x_c$ and $y_c$ are finger center, $a$ and $b$ are the semi-major and semi-minor axes of the finger region. 
The rolling transform applies an angular offset 
\begin{equation}
\Delta\theta(u,v) = \psi \cdot \sqrt{u^2+v^2},
\end{equation}
where $\psi \sim U(-\psi_{\max},\psi_{\max})$ controls the maximum rolling degree. 
The transformed coordinates are mapped as
\begin{equation}
I^\prime(x',y') = R_{\Delta\theta(u,v)}(x,y),
\end{equation}
with $R_{\Delta\theta}$ denoting a local rotation operator centered at $(x_c,y_c)$.

\noindent\textbf{Scaling:} $T_{\text{scale}}(I;s)$ uniformly scales fingers by a scaling factor $s$.

\noindent\subsubsection{\textbf{\textit{Optical Degradation (Grayscale/Render Stage)}}}

\noindent\textbf{Over/Under Exposure:} $T_{\text{expo}}(I; (c_x,c_y),v_0)$ introduces local exposure through a centrally symmetric, exponentially decaying exposure field. 
$(c_x,c_y)$ is a random center sampled within the finger region and $d(x,y)=\lVert(x,y)-(c_x,c_y)\rVert_2$. Denote $d_{\max}=\max_{(x,y)\in\mathrm{ROI}} d(x,y)$. $v_0$ denotes the maximum exposure exponent. The exposure exponent field is
\begin{equation}
e(x,y)=v_0 - \delta\, d(x,y),\, \delta=\frac{v_0}{d_{\max}},
\end{equation}
and the multiplicative gain is $f(x,y)=2^{e(x,y)}$. The operator applies
\begin{equation}
I'(x,y)=\operatorname{clip}\big( f(x,y)\,I(x,y),\;0,\;255 \big).
\end{equation}

\noindent\textbf{Skin Scattering:}
$T_{\text{scatter}}(I;\alpha,\beta,A_{bm})$ employs an atmospheric scattering model to simulate skin subsurface scattering and haze effects in the near-infrared spectrum.

The spatially varying transmittance is defined as
\begin{equation}
t(x,y)=\exp\!\big(-\beta\, d_s(x,y)\big),
\end{equation}
and the scattered image is computed by
\begin{equation}
I'(x,y)=I(x,y)\,t(x,y) + A_{bm}\big(1-t(x,y)\big),
\end{equation}

where $d_s(x,y)$ is a spatial distance proxy 
(e.g., $d_s(x,y)=\max(0,-\alpha\cdot\mathrm{dist}(x,y)+C)$),
$\alpha$ is a spatial decay factor, $\beta$ is the scattering coefficient,
and $A_{bm}$ denotes the ambient light~\cite{nayar1999vision}.

\noindent\textbf{Motion Blur:} $T_{\text{motion}}(I;L,\theta)$ simulates relative motion by convolving with a direction-dependent linear kernel $K_{L,\theta}$:
Motion blur is simulated by convolution with a normalized linear kernel $K_{L,\theta}$:
\begin{equation}
I'(x,y)=(I(x,y) * K_{L,\theta})(x,y),
\end{equation}
where $K_{L,\theta}$ denotes a discrete line segment kernel of length $L$ along direction $\theta$, normalized.

\noindent\textbf{Optical Blur:} $T_{\text{opt}}(I;\sigma)$ approximates the defocus or residual blur in an imaging system using a Gaussian point spread function (PSF):
\begin{equation}
I'(x,y)=(I * G_{\sigma})(x,y),\, G_{\sigma}(r)=\mathcal{N}(0,\sigma^2),
\end{equation}

\subsubsection{\textbf{\textit{{Triggering Strategy}}}}

The ICDG adopts a two-stage probabilistic operator triggering strategy. Specifically, for each sample, the number of activated operators is first drawn, and then a subset of operators is selected from the operator family according to normalized weights without replacement.

Formally, the activation tendency of each operator $k \in \mathcal{P}$ is encoded by a non-negative base weight $\omega_k > 0$, which is normalized into a probability $p_k = \frac{\omega_k}{\sum_{j \in \mathcal{P}} \omega_j}.$
The number of active operators is modeled as a Poisson random variable $K \sim \mathrm{Poisson}(\lambda)$,
truncated $K \le K_{\max}$. This yields a long-tailed distribution, where most samples undergo only a few perturbations, while a small number of samples are subject to stronger distortions. This better reflects the intra-class variability observed in real finger vein imagery.

Finally, $K$ operators are sampled without replacement from $\mathcal{P}$ according to $\{p_k\}$, forming the activated set $\mathcal{P}_{\mathrm{act}}$.

\begin{figure*}[t]
	\centering
	\includegraphics[scale=1,width=0.98\textwidth]{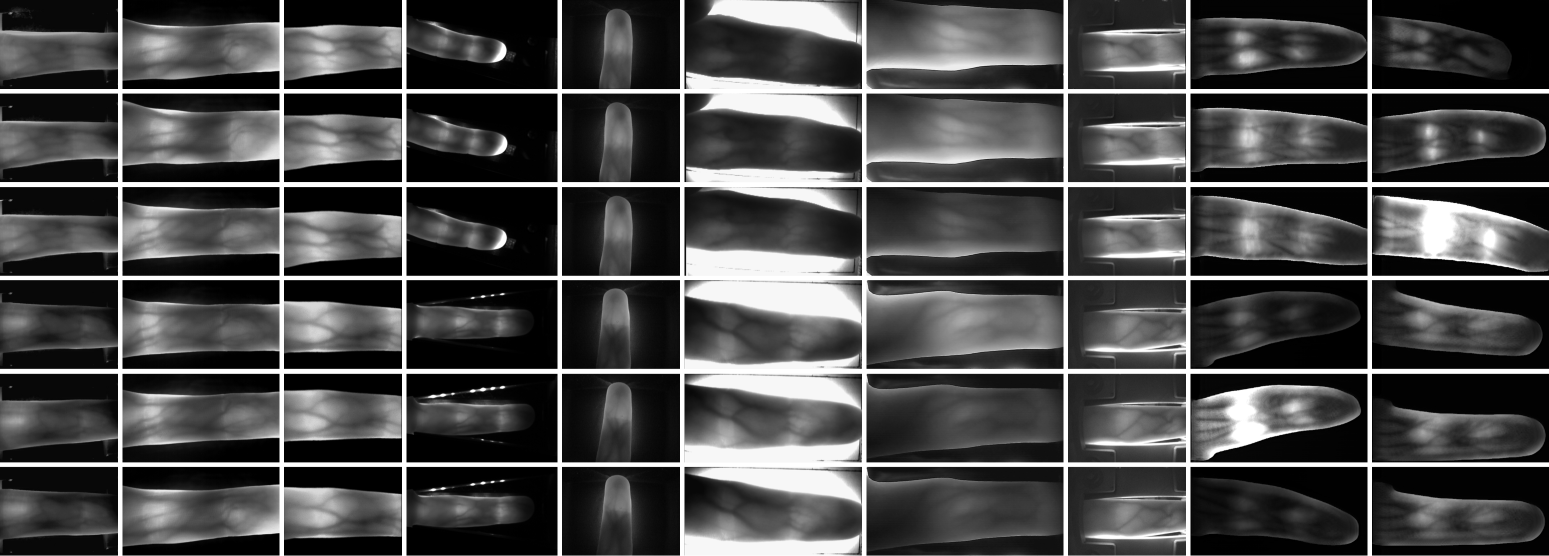}
	\caption{Comparison between real and synthetic finger vein images generated by FVeinSyn. From left to right: SDUMLA-FV, UTFVP, MMCBNU\_6000, PLUS-FV3, FV-USM, HKPU-FV, SCUT-FV3, THU-FV3, and FVeinSyn generate synthetic finger vein images (two columns).}
	\label{fig4}
\end{figure*}

\begin{figure}[t]
	\centering
	\includegraphics[scale=1,width=0.45\textwidth]{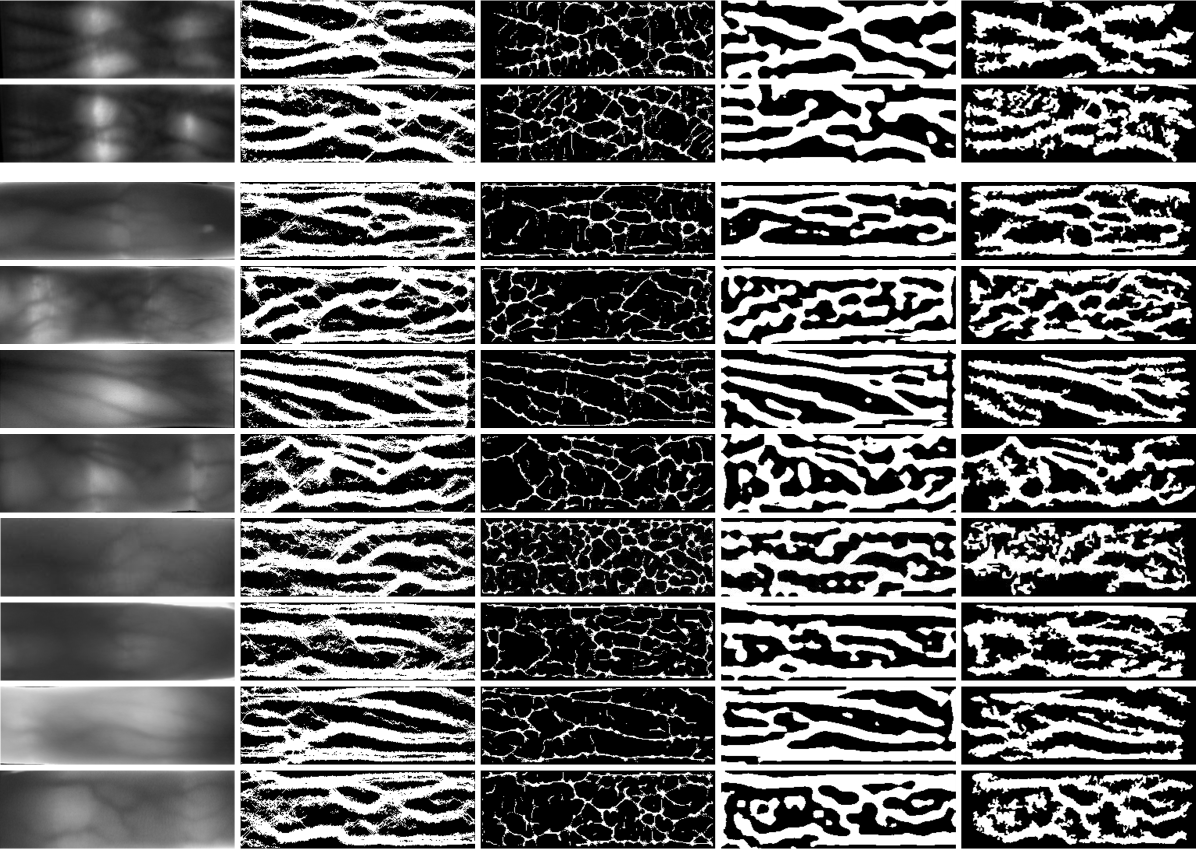}
	\caption{Vein pattern extraction results on real and synthetic finger vein images using classical methods, including RLT, MC, Gabor filter, and IUWT. From top to bottom: FVeinSyn-generated synthetic finger vein images (two rows), SDUMLA-FV, UTFVP, MMCBNU\_6000, PLUS-FV3, FV-USM, HKPU-FV, SCUT-FV3, and THU-FV3.}
	\label{fig5}
\end{figure}

\section{Experiment Results}
In this section, we conduct a comprehensive evaluation of the proposed FVeinSyn framework. We first assess the realism and quality distribution of generated images. Subsequently, quantitative recognition experiments are conducted to compare the performance of real and synthetic data under various scenarios, including open-set recognition, cross-domain recognition, and sample/identity-limited recognition. Finally, we analyze the synthetic dataset on identity uniqueness, vascular patterns consistency, and intra-class diversity.

\subsection{Experimental Setup}
\noindent
\textit{\textbf{Datasets}}:
We utilized eight publicly available datasets: SDUMLA-FV~\cite{yin2011sdumla}, UTFVP~\cite{Ton_UTFVPDB2013}, MMCBNU\_6000~\cite{lu2013available}, PLUS-FV3~\cite{8698588}, FV-USM~\cite{asaari2014fusion}, HKPU-FV~\cite{kumar2011human}, SCUT-FV3~\cite{tang2019finger}, and THU-FV3~\cite{6467066} totaling 3938 identities and 35588 finger vein samples. During finger vein sample generation, since not all datasets contained complete finger regions, only HKPU-FV, FV-USM, and PLUS-FV3 were used for region supervision and shape supervision.

\noindent
\textit{\textbf{Generator Model Training Configuration}}:
We generated 10,000 identities, each with 50 samples. The synthetic vein images had dimensions of $300\times600$, while the finger-shaped vein images and finger vein ROI images were sized at $100\times300$. We employed the Adam optimizer ($\beta_1=0.5$, $\beta_2=0.999$) for model training, with an initial learning rate of $2e-4$ that decayed after 50 epochs. The generator model was implemented using the PyTorch framework and trained on 4 NVIDIA RTX 4090 GPUs with a batch size of 24.

\noindent
\textit{\textbf{Recognition Model Training Configuration}}:
The recognition model utilized a ResNet101 backbone~\cite{he2016deep} integrated with ArcFace~\cite{deng2019arcface} (scale factor $s=64$, margin $M=0.5$). The recognition model is
firstly pretrained on synthesized data for 20 epochs and then
finetuned on real datasets for 50 epochs. Finger vein ROI images sized at $100 \times 300$, with a learning rate of $0.1$ for training and $1e-2$ for fine-tuning, respectively. The model was implemented in PyTorch and trained on 4 NVIDIA RTX 4090 GPUs with a batch size of $256$.

\subsection{Visualization Analysis}
To visually demonstrate the effectiveness of the FVeinSyn framework, we first present the generated images. Fig. \ref{fig4} shows a comparison between real samples (the first eight columns) and FVeinSyn-generated synthetic samples (the last two columns), with each column containing three rows corresponding to different samples of the same subject. As shown, the synthetic samples exhibit a high degree of anatomical consistency: the main veins are clearly delineated, branching structures are coherent, and the overall topology is anatomically plausible. In terms of appearance, the synthetic images closely resemble real near-infrared acquisitions, exhibiting similar local illumination, scattering, and pattern details.

To further validate the similarity of the synthetic images at the pattern level, several classical vein pattern feature extraction methods were applied to both real and synthetic samples, including repeated line tracking~\cite{RLT}, maximum curvature~~\cite{MC}, Gabor filter~\cite{kumar2011human}, and isotropic undecimated wavelet transform~\cite{IUWT}. Fig.~\ref{fig5} presents the extraction results of these methods on both real and synthetic images. It can be observed that, in terms of pattern locations, line connectivity, and local response intensity, the pattern feature distributions of the synthetic images closely match those of the real images, indicating that FVeinSyn effectively preserves vein pattern structures and local details.

\subsection{Image Quality Assessment}
We further conducted a quantitative assessment of image quality across eight publicly available real finger vein datasets and three synthetic datasets. The experiments employed five reference-free image quality metrics: global contrast (GCF)~\cite{matkovic2005global}, Entropy~\cite{tsai2008information}, Tenengrad, natural image quality evaluator (NIQE)~\cite{mittal2012making}, and blind image space quality evaluator (BRISQUE)~\cite{moorthy2011blind}. Among these, GCF reflects the separability between the vein pattern region and surrounding tissue, entropy characterises the information richness of the vein pattern, and the Tenengrad gradient is used to assess the sharpness of the generated image. NIQE and BRISQUE evaluate the overall visual quality of the image from the perspectives of naturalness and perceptual distortion, respectively.

\begin{figure}[t]
	\centering
	\includegraphics[scale=1,width=0.48\textwidth]{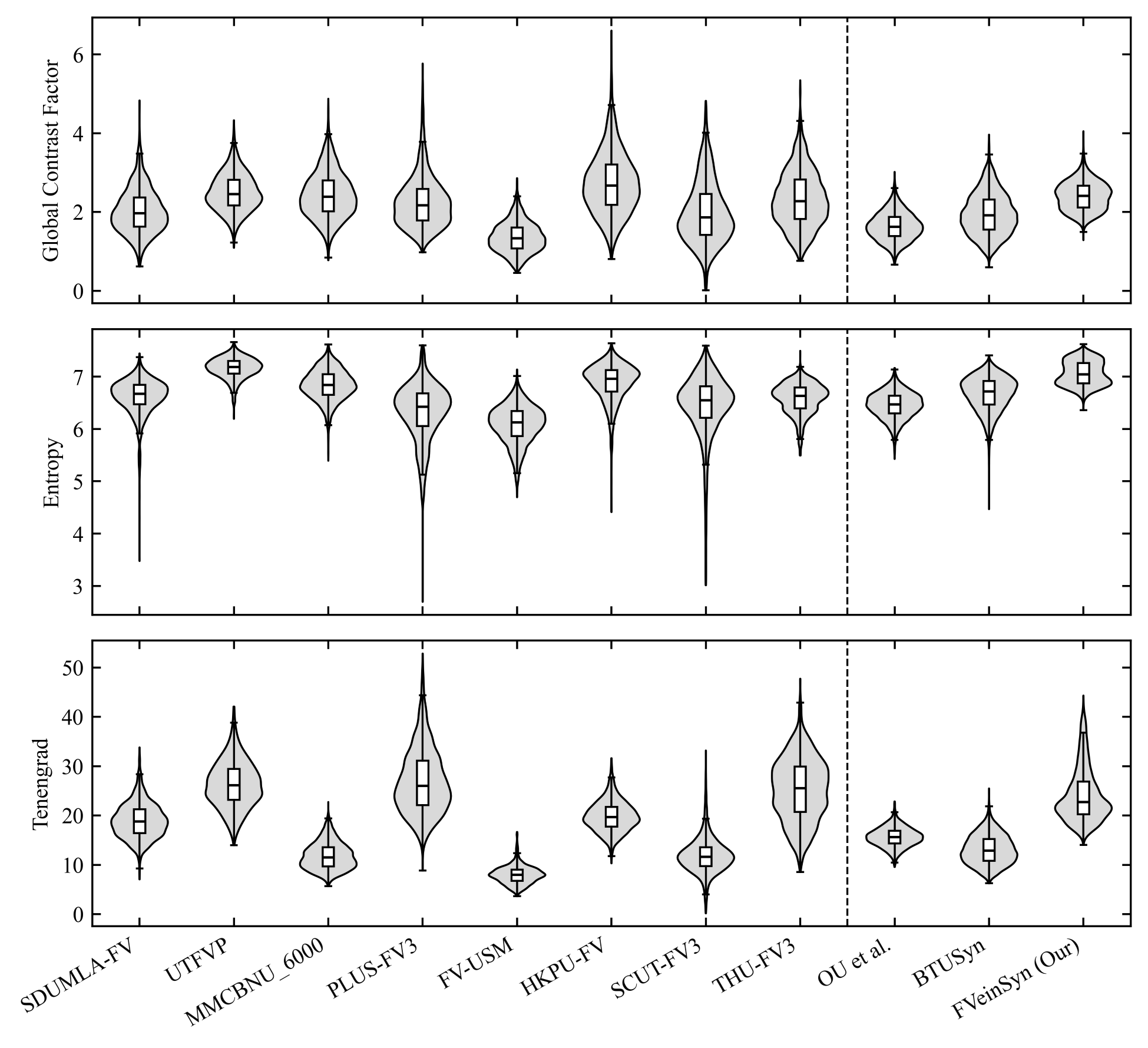}
	\caption{Global contrast factor, Entropy, and Tenengrad gradient distributions of image quality metrics for real and synthetic finger vein datasets.}
	\label{fig6}
\end{figure}

Table \ref{tb2} lists the average scores across the five metrics for each dataset. Fig. \ref{fig6} further illustrates the score distributions for real and synthetic datasets across the three key metrics: GCF, entropy, and Tenengrad. The interpretation of these metrics is closely related to their underlying physical meanings and the characteristics of real datasets. Specifically, GCF measures global image contrast and reflects the discriminability between vein patterns and surrounding tissues. Higher GCF values indicate clearer and more distinguishable vein structures. 

As shown in Fig.~\ref{fig5}, the UTFVP and MMCBNU\_6000 datasets exhibit relatively clear vein structural characteristics, and accordingly present higher GCF values. In comparison, the GCF distribution of images generated by FVeinSyn is closer to that of these high-quality real datasets. 
Similarly, entropy reflects the information representation characteristics and statistical complexity of vein patterns within the region of interest. 
In general, clearer vein structures are associated with higher entropy values. 
From this perspective, the entropy distribution of FVeinSyn-generated images is also highly consistent with those of the UTFVP and MMCBNU\_6000 datasets. 
In addition, Tenengrad measures image sharpness by quantifying gradient strength, where higher values indicate clearer structural edges and richer details. 
Taken together, these results show that the proposed FVeinSyn method achieves statistical distributions on key metrics, including GCF, entropy, and Tenengrad gradient, that are most closely aligned with high-quality real datasets, while exhibiting overall superior performance compared to other synthetic approaches. 
This suggests that FVeinSyn-generated finger vein images demonstrate high authenticity and naturalness in terms of contrast, information representation, and clarity, thereby validating the effectiveness of the proposed method in modeling realistic pattern and overall visual quality.

\begin{table*}[t]
	\centering
	\caption{Comparison of image quality metrics across different finger vein datasets.}
	\label{tb2}
	\resizebox{\textwidth}{!}{
		\setlength{\tabcolsep}{2pt}
		\rowcolors{2}{gray!10}{white}
		\begin{tabular}{l*{9}{S[table-format=2.2]}|*{3}{S[table-format=2.2]}}
			\toprule
			\multirow{2}{*}{Metrics} & \multicolumn{9}{c|}{Real Datasets} 
			& \multicolumn{3}{c}{Synthetic Datasets} \\
			\cmidrule(lr){2-10} \cmidrule(lr){11-13}
			& {SDUMLA-FV} & {UTFVP} & {MMCBNU\_6000} & {PLUS-FV3} & {FV-USM} & {HKPU-FV} & {SCUT-FV} & {THU-FV3} & {Avg.}
			& {Ou \emph{et al.}~\cite{ou2022gan}} & {BTUSyn~\cite{hillerstrom2014generating}} & {FVeinSyn(Our)} \\
			\midrule
			GCF       & 2.02 & 2.50 & 2.43 & 2.26 & 1.35 & 2.73 & 1.99 & 2.33 & 2.16 & 1.64 & 1.95 & 2.40 \\
			Entropy   & 6.62 & 7.16 & 6.84 & 6.33 & 6.08 & 6.88 & 6.44 & 6.58 & 6.64 & 6.45 & 6.67 & 7.06 \\
			Tenengrad & 18.96 & 26.31 & 11.79 & 26.92 & 7.96 & 19.83 & 11.87 & 25.26 & 18.73 & 15.61 & 13.09 & 24.12 \\
			NIQE      & 30.28 & 31.18 & 32.74 & 24.84 & 31.86 & 31.53 & 31.53 & 33.27 & 31.16 & 27.02 & 29.90 & 28.59 \\
			BRISQUE   & 53.11 & 46.09 & 67.52 & 57.56 & 61.46 & 67.08 & 69.00 & 35.89 & 56.94 & 58.94 & 88.37 & 44.72 \\
			\bottomrule
		\end{tabular}
	}
\end{table*}

\begin{table}[t]
	\centering
	\setlength{\tabcolsep}{3pt} 
	\caption{Average realism ratings of different synthesis methods (9 = Most Realistic, 1 = Least Realistic)}
	\label{tb3}
	\begin{tabular}{llS[table-format=1.2]S[table-format=1.2]}
		\toprule
		Method & Region & {Avg. Realism Rating} & {Std. Dev.} \\
		\midrule
		BTUSyn~\cite{hillerstrom2014generating} & ROI & 3.74 & 0.60 \\
		Ou \emph{et al.}~\cite{ou2022gan} & ROI & 4.61 & 0.78 \\
		FVeinSyn (Our) & ROI & 5.65 & 0.53 \\
		FVeinSyn (Our) & Full Finger & 7.11 & 0.56 \\
		\bottomrule
	\end{tabular}
\end{table}

\begin{table}[t]
	\centering
	\setlength{\tabcolsep}{3pt} 
	\caption{percentage classified as real per synthesis data}
	\label{tb4}
	\begin{tabular}{llS[table-format=2.2]S[table-format=2.2]}
		\toprule
		Method & Region & {Average Rating (\%)} & {Std. Dev. (\%)} \\
		\midrule
		BTUSyn~\cite{hillerstrom2014generating} & ROI & 20.50 & 14.80 \\
		Ou \emph{et al.}~\cite{ou2022gan} & ROI & 40.84 & 16.23 \\
		FVeinSyn (Our)& ROI & 82.97 & 10.98 \\
		FVeinSyn (Our)& Full Finger & 90.54 & 8.03 \\
		\bottomrule
	\end{tabular}
\end{table}

\begin{table*}[t]
	\centering
	\setlength{\tabcolsep}{3pt} 
	\caption{Performance comparison of various enhancement or synthesis methods for finger vein recognition}
	\label{tb5}
	\rowcolors{2}{gray!10}{white}
	\begin{tabular}{lccc*{9}{S[table-format=1.4]}}
		\toprule
		\multirow{2}{*}{Method} & \multicolumn{3}{c}{Configs} & \multicolumn{9}{c}{Performance (TAR@FAR=1e-3) $\uparrow$} \\ 
		\cmidrule(lr){2-4} \cmidrule(lr){5-13}
		& \#IDs & \#per ID & \#Image & {SDUMLA-FV} & {UTFVP} & {MMCBNU\_6000} & {PLUS-FV3} & {FV-USM} & {HKPU-FV} & {SCUT-FV3} & {THU-FV3} & {Avg.} \\ 
		\midrule
		MC & - & - & - & 0.3117 & 0.5593 & 0.7669 & 0.6895 & 0.7632 & 0.3066 & 0.6261 & 0.5016 & 0.5656 \\
		RLT & - & - & - & 0.4006 & 0.7093 & 0.7650 & 0.5363 & 0.8406 & 0.4753 & 0.5853 & 0.5788 & 0.6114 \\
		IUWT & - & - & - & 0.4273 & 0.8269 & 0.7951 & 0.6190 & 0.7955 & 0.5758 & 0.5775 & 0.5405 & 0.6447 \\
		GF & - & - & - & 0.3820 & 0.7232 & 0.8640 & 0.5104 & 0.8585 & 0.4743 & 0.5721 & 0.6158 & 0.6250 \\ 
		\midrule
		Real data & 1.9k & - & 18k & 0.4602 & 0.7593 & 0.8385 & 0.5385 & 0.7709 & 0.6447 & 0.5504 & 0.5799 & 0.6428 \\
		Ou \emph{et al.} & 3k & 1 & 3k & 0.5692 & 0.6852 & 0.8239 & 0.5004 & 0.7340 & 0.6223 & 0.5283 & 0.5118 & 0.6219 \\
		BTUSyn & 1.6k & 10 & 16k & 0.6262 & 0.8222 & 0.8383 & 0.4614 & 0.7856 & 0.6938 & 0.5478 & 0.5772 & 0.6691 \\
		FVeinSyn & 1.9k & 10 & 19k & 0.7155 & 0.9222 & 0.9169 & 0.6852 & 0.8859 & 0.8903 & 0.5729 & 0.7922 & 0.7976 \\
		FVeinSyn & 1.9k & 50 & 95k & 0.7941 & 0.9546 & 0.9438 & 0.8156 & 0.9305 & 0.9203 & 0.6532 & 0.8875 & 0.8625 \\
		FVeinSyn & 10k & 10 & 100k & 0.8031 & 0.9500 & 0.9457 & 0.7935 & 0.9098 & 0.9295 & 0.6574 & 0.8714 & 0.8576 \\
		FVeinSyn & 10k & 50 & 500k & \textbf{0.9402} & \textbf{0.9778} & \textbf{0.9538} & \textbf{0.9132} & \textbf{0.9551} & \textbf{0.9655} & \textbf{0.7078} & \textbf{0.9233} & \textbf{0.9171} \\
		\bottomrule
	\end{tabular}
\end{table*}

\begin{figure*}[t]
	\centering
	\includegraphics[scale=1,width=0.95\textwidth]{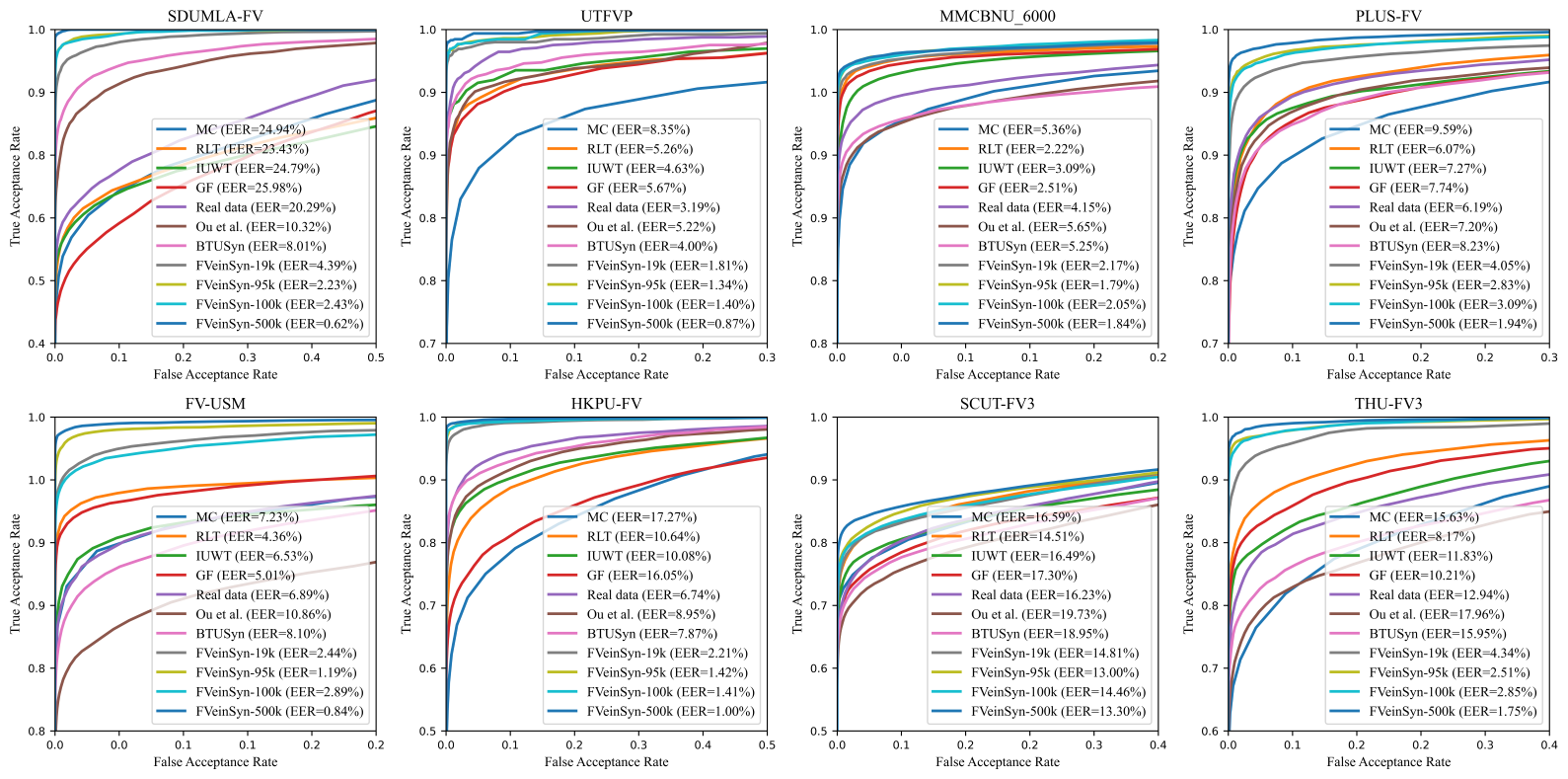}
	\caption{Comparison of receiver operating characteristic (ROC) curves and equal error rates (EERs) for FVeinSyn and other representative methods under the open-set finger vein recognition protocol.}
	\label{fig7}
\end{figure*}

\subsection{CrowdSourced Evaluation}
To assess the realism of the samples generated by FVeinSyn, we conducted a crowdsourced study involving ten researchers specializing in finger vein recognition. The evaluation focused on two aspects: (1) comparing the perceptual realism of our synthetic images with real data and several representative synthesis methods, and (2) measuring the probability of synthetic images were perceived as real. Evaluation samples were drawn from eight public datasets, two baseline synthesis approaches (Ou \emph{et al.}~\cite{ou2022gan} and BTUSyn~\cite{hillerstrom2014generating}), and the proposed FVeinSyn framework. Both full-finger and ROI image types were assessed independently.

For the realism scoring task, observers were shown groups of nine images containing a mix of real and synthesized samples. Each image was rated on a 1–9 scale, where 9 indicates the highest realism. Real and synthetic samples were balanced within each group, and both the order and composition were randomized. Each participant evaluated at least 20 groups. Table~\ref{tb3} summarizes the mean scores and standard deviations. FVeinSyn obtained the highest realism ratings for both full-finger and ROI images. Traditional modeling-based synthesis (BTUSyn~\cite{hillerstrom2014generating}) exhibited noticeably lower realism, while learning-based methods (Ou \emph{et al.}~\cite{ou2022gan}) produced improved results but still suffered from local artifacts arising from limited supervision.

In the real–fake discrimination task, observers were shown at least 90 images in random order and asked to decide whether each was “real’’ or “synthetic.’’ Table~\ref{tb4} reports the proportion of images from each method that were judged as real. FVeinSyn again achieved the highest perceived-real rate, reaching 90.54~$\pm$~8.03\% for full-finger images and 82.97~$\pm$~10.98\% for ROI images. These results collectively demonstrate that FVeinSyn produces synthetic samples with perceptual realism and naturalness closest to real data.

\subsection{Recognition Performance Evaluation}

\noindent\textbf{\textit{Open-Set Finger Vein Recognition:}} Under the open-set protocol, we conducted a systematic comparison between real and synthetic data for finger vein recognition. The experiments were designed with a strict non-overlapping identity partition, ensuring that the vein identities in the training and test sets are completely independent. We compared the proposed FVeinSyn with representative traditional pattern-based matching methods~\cite{MC, RLT, IUWT, kumar2011human, hillerstrom2014generating} and learning–based approaches~\cite{ou2022gan}. The recognition performance across multiple datasets is summarized in Table~\ref{tb5}. The results demonstrate that synthetic finger vein images generated by FVeinSyn exhibit a clear advantage: under the same number of identities and sample scale, the recognition performance improves by an average of 15.48\% compared with models trained on real data. When the training set size and sample diversity are further increased, the improvement reaches up to 27.43\%. Compared with other finger vein synthesis methods, FVeinSyn consistently achieves higher recognition accuracy and more stable performance across all datasets. Fig.~\ref{fig7} presents the ROC curves and equal error rates (EERs) of different methods on various datasets, where FVeinSyn attains the largest area under the curve (AUC) and the lowest EER. Results show that FVeinSyn effectively improving the performance and robustness of recognition systems trained with synthetic data.

\noindent
\textbf{\textit{Cross-Domain Recognition:}} Cross-domain performance remains a critical challenge for practical finger vein recognition systems. We evaluated models trained with either real or synthetic data and tested their performance across different domains. The quantitative results are summarized in Table~\ref{tb6}. Experimental findings show that pretraining with FVeinSyn-generated data significantly enhances cross-domain generalization, achieving an average recognition improvement of 40.87\% per dataset and over 50\% on several benchmarks. The largest improvement, 61.44\%, is observed on the HKPU-FV dataset, which exhibits notable differences from other datasets, such as brighter backgrounds and distinct finger regions.  
The poor cross-domain performance of models trained solely on real data primarily stems from limited dataset size and domain bias introduced by varying imaging conditions. These constraints hinder the model’s ability to generalize across variations in illumination, background, and finger positioning. By synthesizing samples with diverse inter-class and cross-domain variations, FVeinSyn effectively addresses this limitation, enabling the model to learn more comprehensive and robust feature representations, thereby achieving substantially improved cross-domain recognition performance.

\begin{table*}[t]
	\centering
	\caption{Cross-dataset evaluation using or not using FVeinSyn for pre-training.
		\CheckmarkBold indicates that pre-training with synthetic data is used (TAR@FAR=1e-3) $\uparrow$}
	\label{tb6}
	\setlength{\tabcolsep}{3pt}
	\resizebox{\textwidth}{!}{
		\begin{tabular}{l|c*{8}{S[table-format=1.4]S[table-format=1.4]}}
			\toprule
			\multicolumn{2}{c}{\diagbox{Train}{Test}} & \multicolumn{2}{c}{SDUMLA\_FV} & \multicolumn{2}{c}{UTFVP} & \multicolumn{2}{c}{MMCBNU\_6000} & \multicolumn{2}{c}{PLUS-FV3} & \multicolumn{2}{c}{FV-USM} & \multicolumn{2}{c}{HKPU-FV} & \multicolumn{2}{c}{SCUT-FV3} & \multicolumn{2}{c}{THU-FV3} \\
			\midrule
			\multirow{2}{*}[1ex]{PT w/ Syn} & & {\XSolidBrush} & {\CheckmarkBold} & {\XSolidBrush} & {\CheckmarkBold} & {\XSolidBrush} & {\CheckmarkBold} & {\XSolidBrush} & {\CheckmarkBold} & {\XSolidBrush} & {\CheckmarkBold} & {\XSolidBrush} & {\CheckmarkBold} & {\XSolidBrush} & {\CheckmarkBold} & {\XSolidBrush} & {\CheckmarkBold} \\
			
			\midrule
			SDUMLA\_FV & & {--} & {--} & 0.4065 & 0.8727 & 0.5565 & 0.9158 & 0.1344 & 0.7164 & 0.3117 & 0.7876 & 0.1317 & 0.7426 & 0.0120 & 0.4293 & 0.4464 & 0.6962 \\
			UTFVP & & 0.5083 & 0.8007 & {--} & {--} & 0.5850 & 0.8953 & 0.1611 & 0.7738 & 0.2421 & 0.7654 & 0.1375 & 0.8118 & 0.0281 & 0.4601 & 0.4834 & 0.6973 \\
			MMCBNU\_6000 & & 0.5137 & 0.8230 & 0.2704 & 0.8889 & {--} & {--} & 0.2095 & 0.7977 & 0.3787 & 0.8672 & 0.2127 & 0.8203 & 0.0334 & 0.5536 & 0.4572 & 0.6615 \\
			PLUS-FV3 & & 0.5362 & 0.8023 & 0.3167 & 0.8819 & 0.4953 & 0.8401 & {--} & {--} & 0.1954 & 0.8375 & 0.0994 & 0.6554 & 0.0024 & 0.2370 & 0.4583 & 0.6193 \\
			FV-USM & & 0.4822 & 0.8019 & 0.2093 & 0.8991 & 0.7940 & 0.9372 & 0.2323 & 0.8375 & {--} & {--} & 0.2284 & 0.8547 & 0.1225 & 0.6056 & 0.4387 & 0.6821 \\
			HKPU-FV & & 0.5973 & 0.8626 & 0.4630 & 0.9509 & 0.7570 & 0.9264 & 0.1874 & 0.8040 & 0.4162 & 0.8040 & {--} & {--} & 0.1538 & 0.5349 & 0.5000 & 0.6699 \\
			SCUT-FV3 & & 0.5438 & 0.7999 & 0.4625 & 0.9560 & 0.7346 & 0.9471 & 0.3159 & 0.6582 & 0.6913 & 0.8983 & 0.2770 & 0.8643 & {--} & {--} & 0.4806 & 0.6787 \\
			THU-FV3 & & 0.5520 & 0.8168 & 0.5264 & 0.7444 & 0.7396 & 0.9355 & 0.2333 & 0.7475 & 0.3318 & 0.8323 & 0.1897 & 0.8066 & 0.0536 & 0.6485 & {--} & {--} \\
			\midrule
			\midrule
			Avg. & &
			0.5334 & 0.8153 &
			0.3793 & 0.8848 &
			0.6660 & 0.9139 &
			0.2106 & 0.7622 &
			0.3667 & 0.8275 &
			0.1823 & 0.7937 &
			0.0580 & 0.4956 &
			0.4664 & 0.6721 \\
			Imp. & &
			\multicolumn{2}{c}{\textbf{\textcolor{darkgreen}{$\nearrow$}} 0.2819} &
			\multicolumn{2}{c}{\textbf{\textcolor{darkgreen}{$\nearrow$}} 0.5056} &
			\multicolumn{2}{c}{\textbf{\textcolor{darkgreen}{$\nearrow$}} 0.2479} &
			\multicolumn{2}{c}{\textbf{\textcolor{darkgreen}{$\nearrow$}} 0.5516} &
			\multicolumn{2}{c}{\textbf{\textcolor{darkgreen}{$\nearrow$}} 0.4608} &
			\multicolumn{2}{c}{\textbf{\textcolor{darkgreen}{$\nearrow$}} 0.6114} &
			\multicolumn{2}{c}{\textbf{\textcolor{darkgreen}{$\nearrow$}} 0.4376} &
			\multicolumn{2}{c}{\textbf{\textcolor{darkgreen}{$\nearrow$}} 0.2058} \\
			\bottomrule
	\end{tabular}}
\end{table*}

\noindent\textbf{\textit{Identify Limited Recognition:}} We consider an identity-limited open-set recognition scenario to compare the performance of different finger vein synthesis methods. Using a 1:1 open-set protocol, we evaluate recognition performance as the number of training identities gradually increases (50, 100, 200, 500, 1000, and 1969), as summarized in Table~\ref{tb8}.

The results show that when the number of training identities is extremely small (50 IDs), both Ou \emph{et al.}~\cite{ou2022gan} and BTUSyn~\cite{hillerstrom2014generating} exhibit limited recognition performance, failing to ensure model stability. In contrast, models pretrained with FVeinSyn-generated images maintain strong recognition performance even under such highly constrained conditions. Notably, when only 100 training identities are used (5\% of the full set), the FVeinSyn-pretrained model still outperforms the baseline model trained on all 1,969 real identities. These findings indicate that FVeinSyn effectively enhances inter-class diversity and discriminative feature learning under limited-identity conditions, thereby improving model generalization and open-set recognition performance.
\begin{table}[]
	\centering
	\setlength{\tabcolsep}{3pt} 
	\caption{Performance comparison with increasing training samples (TAR@FAR=1e-3) $\uparrow$}
	\label{tb8}
	\begin{tabular}{l*{6}{S[table-format=1.4]}}
		\toprule
		\multicolumn{1}{c}{\diagbox{Method}{\#IDs}} & {50} & {100} & {200} & {500} & {1000} & {1969} \\
		\midrule
		Real Data & 0.0008 & 0.1765 & 0.4071 & 0.4926 & 0.5577 & 0.6222 \\
		Ou \emph{et al.}~\cite{ou2022gan} & 0.0870 & 0.3883 & 0.4741 & 0.5310 & 0.5726 & 0.6038 \\
		BTUSyn~\cite{hillerstrom2014generating} & 0.1753 & 0.3890 & 0.4801 & 0.5296 & 0.5775 & 0.6487 \\
		FVeinSyn (Our) & \textbf{0.3148} & \textbf{0.6924} & \textbf{0.8095} & \textbf{0.8142} & \textbf{0.8600} & \textbf{0.8748} \\
		\bottomrule
	\end{tabular}
\end{table}

\begin{table*}[t]
	\centering
	\setlength{\tabcolsep}{2pt} 
	\caption{Performance comparison of one/few-shot finger vein recognition (TAR@FAR=1e-3) $\uparrow$}
	\label{tb7}
	\renewcommand{\arraystretch}{0.9}
	
	\begin{tabular}{c|l*{8}{S[table-format=1.4]}}
		\toprule
		\multicolumn{2}{c}{Dataset} & {SDUMLA-FV} & {UTFVP} & {MMCBNU\_6000} & {PLUS-FV3} & {FV-USM} & {HKPU-FV} & {SCUT-FV3} & {THU-FV3} \\
		\midrule
		
		\multirow{4}{*}{$N=1$} 
		& \cellcolor{gray!10}Real data & \cellcolor{gray!10}0.4736 & \cellcolor{gray!10}0.2815 & \cellcolor{gray!10}\underline{0.6874} & \cellcolor{gray!10}0.2655 & \cellcolor{gray!10}0.4365 & \cellcolor{gray!10}0.2359 & \cellcolor{gray!10}\underline{0.1034} & \cellcolor{gray!10}0.4902 \\
		& Ou \emph{et al.}~\cite{ou2022gan} & 0.5319 & 0.3380 & 0.7338 & \underline{0.3928} & 0.4459 & 0.2649 & 0.0576 & 0.5628 \\
		& \cellcolor{gray!10}BTUSyn~\cite{hillerstrom2014generating} & \cellcolor{gray!10}\underline{0.5855} & \cellcolor{gray!10}\underline{0.5333} & \cellcolor{gray!10}0.8376 & \cellcolor{gray!10}0.3834 & \cellcolor{gray!10}\underline{0.6125} & \cellcolor{gray!10}\underline{0.3849} & \cellcolor{gray!10}0.1023 & \cellcolor{gray!10}\underline{0.6213} \\
		& FVeinSyn (Our) & {\textbf{0.7220}} & {\textbf{0.7315}} & {\textbf{0.8710}} & {\textbf{0.8874}} & {\textbf{0.5650}} & {\textbf{0.8096}} & {\textbf{0.2626}} & {\textbf{0.5826}} \\
		\midrule
		
		\multirow{4}{*}{$N=2$} 
		& \cellcolor{gray!10}Real data & \cellcolor{gray!10}0.5440 & \cellcolor{gray!10}0.4333 & \cellcolor{gray!10}0.7815 & \cellcolor{gray!10}0.3823 & \cellcolor{gray!10}0.6490 & \cellcolor{gray!10}0.3889 & \cellcolor{gray!10}0.4070 & \cellcolor{gray!10}0.5681 \\
		& Ou \emph{et al.}~\cite{ou2022gan} & 0.5810 & 0.4583 & 0.8184 & 0.5367 & 0.5384 & 0.3462 & 0.3030 & 0.6123 \\
		& \cellcolor{gray!10}BTUSyn~\cite{hillerstrom2014generating} & \cellcolor{gray!10}\underline{0.6475} & \cellcolor{gray!10}\underline{0.7139} & \cellcolor{gray!10}\underline{0.8828} & \cellcolor{gray!10}\underline{0.5216} & \cellcolor{gray!10}\underline{0.6930} & \cellcolor{gray!10}\underline{0.4723} & \cellcolor{gray!10}\underline{0.3722} & \cellcolor{gray!10}\underline{0.6973} \\
		& FVeinSyn (Our) & {\textbf{0.8913}} & {\textbf{0.9194}} & {\textbf{0.9645}} & {\textbf{0.9590}} & {\textbf{0.9392}} & {\textbf{0.9643}} & {\textbf{0.7491}} & {\textbf{0.9394}} \\
		\midrule
		
		\multirow{4}{*}{$N=3$} 
		& \cellcolor{gray!10}Real data & \cellcolor{gray!10}0.5975 & \cellcolor{gray!10}{--} & \cellcolor{gray!10}0.7784 & \cellcolor{gray!10}0.5318 & \cellcolor{gray!10}0.7231 & \cellcolor{gray!10}0.5196 & \cellcolor{gray!10}0.5061 & \cellcolor{gray!10}0.6202 \\
		& Ou \emph{et al.}~\cite{ou2022gan} & 0.6641 & {--} & 0.8513 & 0.6078 & 0.6456 & 0.4418 & 0.3554 & 0.7621 \\
		& \cellcolor{gray!10}BTUSyn~\cite{hillerstrom2014generating} & \cellcolor{gray!10}\underline{0.7112} & \cellcolor{gray!10}{--} & \cellcolor{gray!10}\underline{0.9020} & \cellcolor{gray!10}\underline{0.6042} & \cellcolor{gray!10}\underline{0.7640} & \cellcolor{gray!10}\underline{0.5826} & \cellcolor{gray!10}\underline{0.4189} & \cellcolor{gray!10}\underline{0.8253} \\
		& FVeinSyn (Our) & {\textbf{0.9497}} & {--} & {\textbf{0.9786}} & {\textbf{0.9878}} & {\textbf{0.9779}} & {\textbf{0.9868}} & {\textbf{0.8835}} & {\textbf{0.9793}} \\
		\bottomrule
	\end{tabular}
\end{table*}
\begin{figure*}[t]
	\centering
	\includegraphics[scale=1,width=0.95\textwidth]{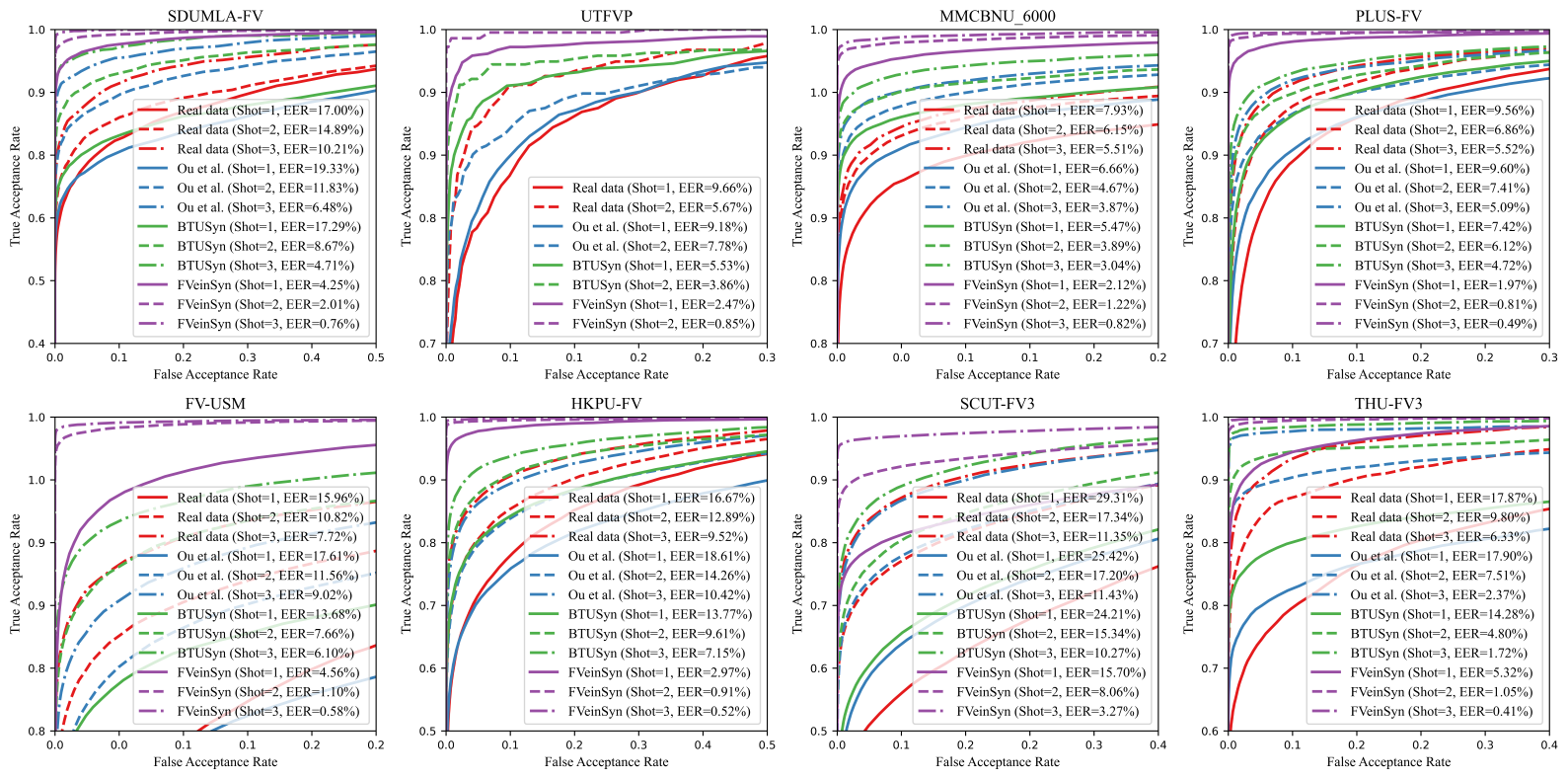}
	\caption{Comparison of receiver operating characteristic (ROC) curves and equal error rates (EERs) for FVeinSyn and other representative methods under the limited finger vein sample recognition protocol.}
	\label{fig8}
\end{figure*}

\noindent\textbf{\textit{Sample Limited (One/Few Shot) Recognition:}} We consider a more practical recognition scenario--sample-limited recognition, users typically provide only a very small number of samples during registration, and sometimes even a single sample. Under this condition, we evaluated recognition models trained with  different synthetic data, and further fine-tuned them with $N$ enrollment samples ($N = 1, 2, 3$). In the experiments, FVeinSyn was systematically compared with real data and two representative finger vein synthesis methods. The results are summarized in Table~\ref{tb7}. Fig.~\ref{fig8} presents the ROC curves and EERs of different sample configurations on various datasets, where FVeinSyn attains the largest area under the curve and the lowest EER.

As shown in the table, models trained solely with a limited number of real samples generally exhibit low recognition performance, making it difficult to ensure system reliability. The introduction of synthetic data alleviates this problem to some extent, with BTUSyn showing moderate improvement over traditional approaches, yet its overall performance remains constrained. In contrast, pretraining with FVeinSyn-generated data followed by fine-tuning with a few real samples leads to a substantial performance improvement. Across all datasets and sample configurations, FVeinSyn consistently achieves the highest recognition accuracy and maintains strong generalization even under the extreme single-sample condition ($N=1$). As the number of enrollment samples increases from $N=1$ to $N=3$, all methods benefit from performance gains, but the improvement obtained by FVeinSyn is the most pronounced. This indicates that the samples synthesized by FVeinSyn better capture inter-class variability and structural consistency, resulting in more representative and learnable features.

\subsection{Synthetic Dataset Evaluation}
In this section, we follow the three dependency-type metrics~\cite{10204758} to evaluate synthetic datasets to help us understand the properties of the generated datasets. We generated 10,000 vein identities using FVeinSyn, with 50 samples per identity, total of 500,000 finger vein images. \( F_{eval} \) is the recognition model used to evaluate synthetic finger vein datasets. The more generalizable \( F_{eval} \) is, the more accurate the metrics are in terms of the identity and diversity of the generated synthetic datasets. Let \( y_c \) be the class label, and \( f_i = F_{eval}(x_i) \). \( d(f_i, f_j) \) represents the distance between two images in the \( F_{\text{eval}} \) feature space.

\noindent
\textbf{\textit{Uniqueness:}}
The uniqueness \(U_{class} \) can be used to quantify the number of distinct identities generated in a synthetic dataset. During evaluation, we determine the capacity limit of the synthetic dataset \(U\) by constructing non-overlapping $r$-ball regions in the \(F_{eval}\) feature space—continuing to add new regions until no further additions can be made without overlap. This process is constrained by two key parameters: the feature distance threshold $r$ (used to determine identity matching) and the feature extraction capability of the \(F_{eval}\) model.

In implementation, we first compute the mean feature vector (cluster center) for all samples belonging to the same identity. If the distance between two cluster centers exceeds threshold \( r \), they are classified as distinct identities. The uniqueness metric \( U_{\text{class}} \) is then defined as \( \frac{|U_c|}{C} \), where the numerator \( |U_c| \) represents the actual count of unique identities generated in the synthetic dataset, and the denominator \( C \) denotes the total number of predefined identity classes in \( F_{eval} \).
As shown in Table \ref{tb9}, the identity distinctiveness of ‌synthetic dataset is ‌99.83\%‌, indicating that the generated finger vein patterns exhibit significant differences. Only a very small portion of the finger vein identity feature vectors show low similarity ($r < 0.2$).

\noindent
\textbf{\textit{Intra-Class Consistency:}}
\(C_{intra}\) is used to evaluate feature coherence among samples belonging to the same unique identity, quantified by measuring the distance between each sample and its corresponding identity feature center. This experiment uses the bounding box annotation provided by the synthetic dataset to extract finger vein ROI images for recognition. Benefiting from the accurate annotation, the synthetic dataset has an intra-class consistency $C_{intra}$ of 100\%.

\begin{table}[t]
	\centering
	\caption{Dataset Evaluation of Synthetic Finger Vein}
	\label{tb9}
	\renewcommand{\arraystretch}{0.5}
	
	\begin{tabular}{l|S[table-format=3.2] S[table-format=3.2] S[table-format=3.2]}
		\toprule
		Dataset & {$U_{class}$} & {$C_{intra}$} & {$D_{intra}$} \\
		\midrule
		Ou \emph{et al.}~\cite{ou2022gan} & 24.57 & {--} & {--} \\
		BTUSyn~\cite{hillerstrom2014generating} & 19.19 & 99.98 & 3.28 \\
		FVeinSyn (Our) & 99.83 & 100.00 & 89.67 \\
		\bottomrule
	\end{tabular}
\end{table}

\begin{figure*}[t]
	\captionsetup[subfloat]{labelsep=none,format=plain,labelformat=empty}
	\centering % 确保整体居中
	
	% 第一行（5图）
	\subfloat[]{
		\includegraphics[width=0.2\linewidth]{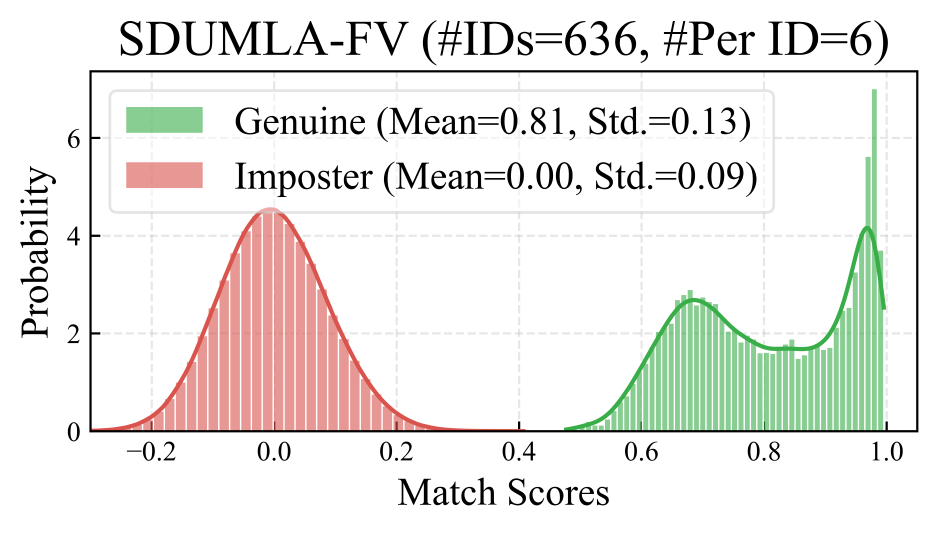}}
	\hspace*{-0.01\linewidth}
	\subfloat[]{
		\includegraphics[width=0.2\linewidth]{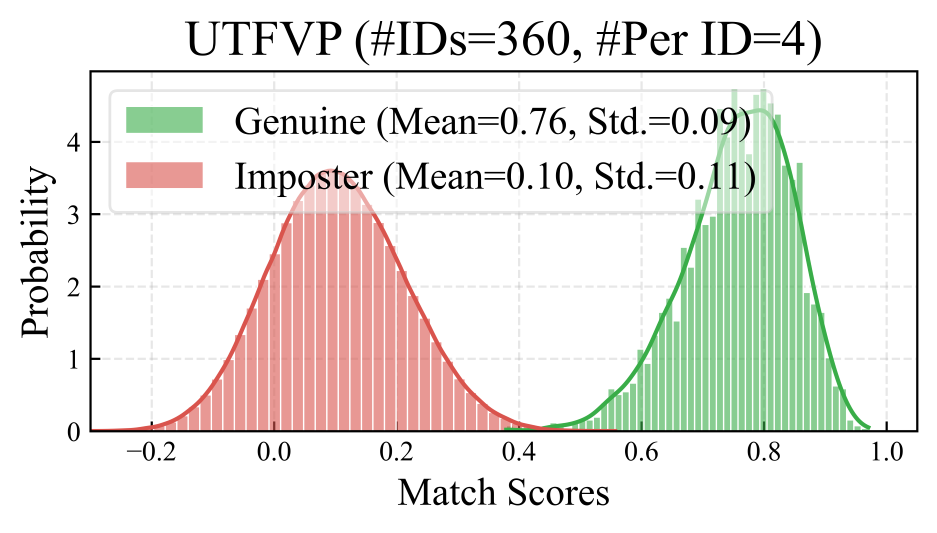}}
	\hspace*{-0.01\linewidth}
	\subfloat[]{
		\includegraphics[width=0.2\linewidth]{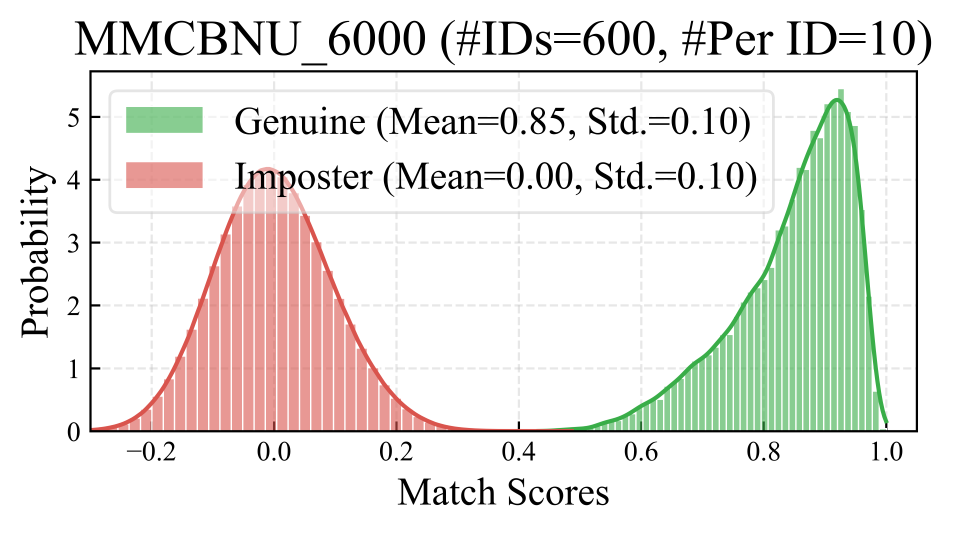}}
	\hspace*{-0.01\linewidth}
	\subfloat[]{
		\includegraphics[width=0.2\linewidth]{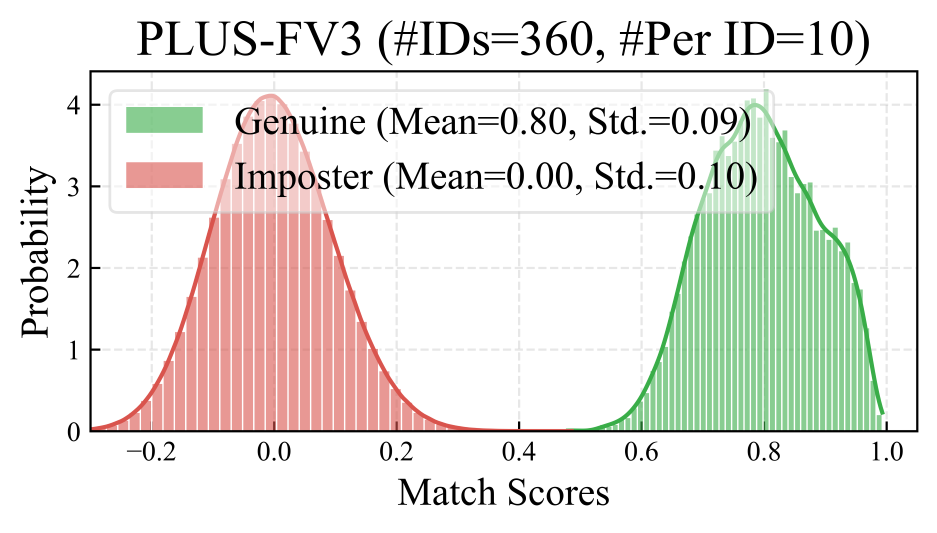}}
	\hspace*{-0.01\linewidth}
	\subfloat[]{
		\includegraphics[width=0.2\linewidth]{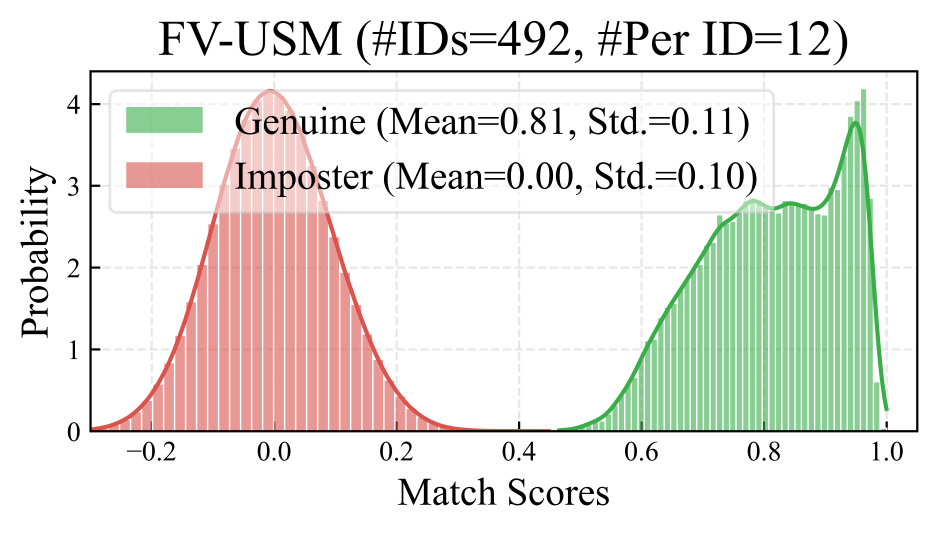}}
	
	\vspace{-28pt} % 调整行间距
	
	% 第二行（5图）
	\subfloat[]{
		\includegraphics[width=0.2\linewidth]{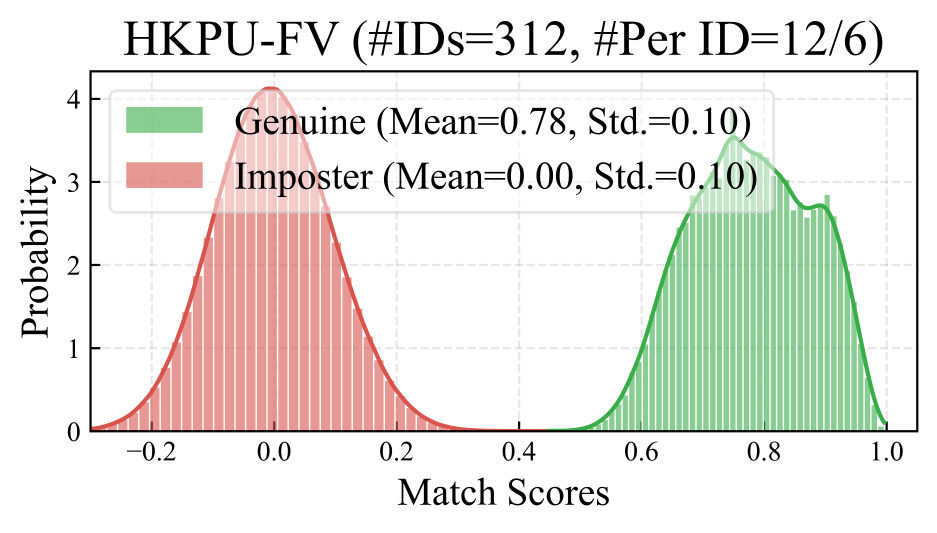}}
	\hspace*{-0.01\linewidth}
	\subfloat[]{
		\includegraphics[width=0.2\linewidth]{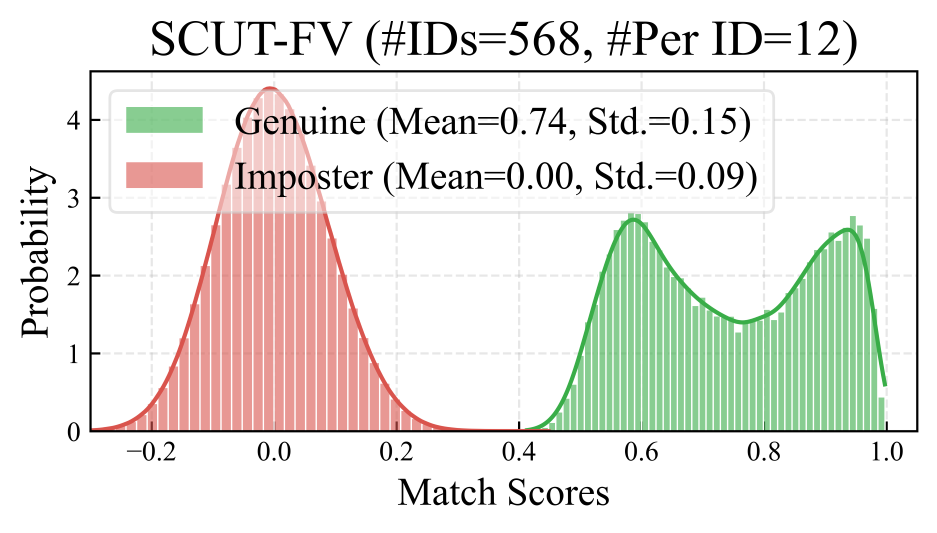}}
	\hspace*{-0.01\linewidth}
	\subfloat[]{
		\includegraphics[width=0.2\linewidth]{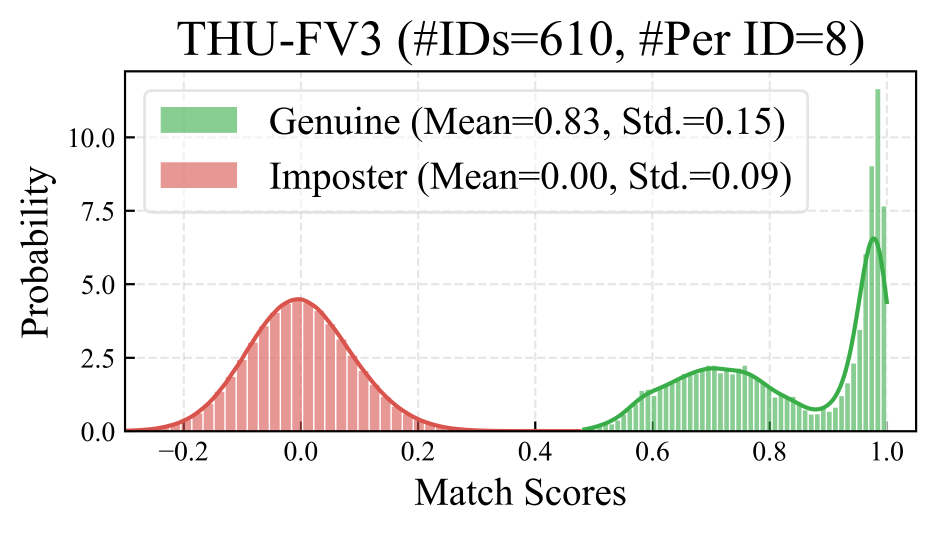}}
	\hspace*{-0.01\linewidth}
	\subfloat[]{
		\includegraphics[width=0.2\linewidth]{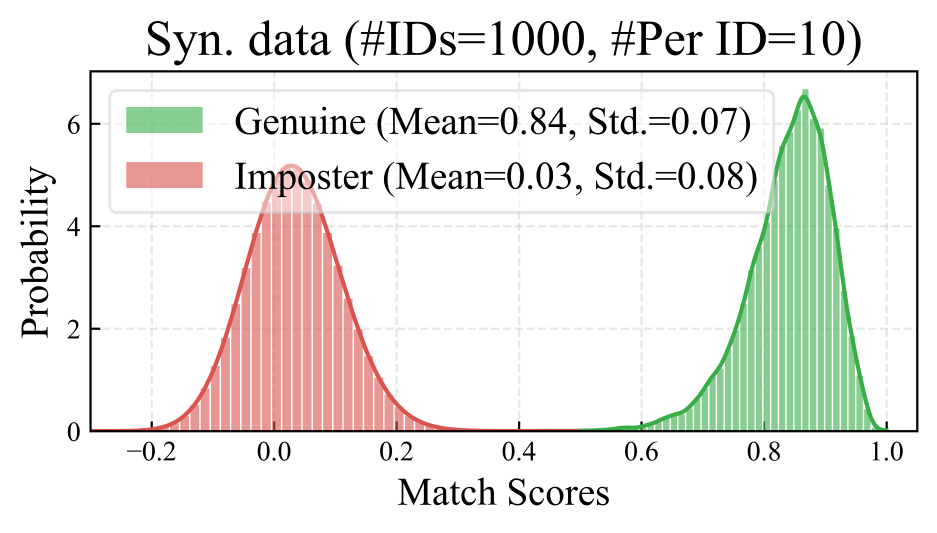}}
	\hspace*{-0.01\linewidth}
	\subfloat[]{
		\includegraphics[width=0.2\linewidth]{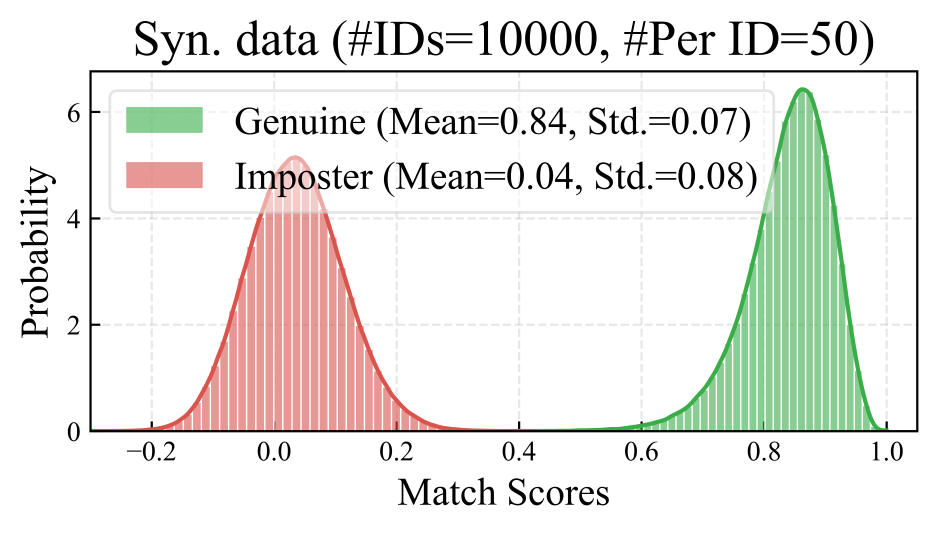}}
	
	\vspace{-10pt}
	
	\caption{Genuine/Impostor matching score distributions across multiple real datasets and the FVeinSyn synthetic dataset.}
	\label{fig9}
\end{figure*}

\noindent
\textbf{\textit{Intra-Class Diversity:}}
‌\(D_{intra}\) measures the variation among generated samples within the same identity. Given the inherently limited intra-class variations in existing public finger vein datasets, we cannot directly adopt the method from~\cite{10204758}. Instead, we train a recognition model \(F_{\mathit{n}}\) using aligned and normalized samples. In this feature space, two samples from the same identity are considered distinct if their distance exceeds threshold $r$. The metric is formally defined as
\begin{equation}
	D_{\text{intra}} = \frac{1}{C}\frac{2}{N(N-1)} \sum_{c=1}^{C} \sum_{i=1}^{N} \sum_{j=i+1}^{N} \delta[d({f_{\mathit{n}}}_i^{c}, {f_{\mathit{n}}}_j^{c}) < r],
\end{equation}

where \( f_{n}^{c} \) is the \( n \)-th sample of class \( c \), and \( \delta[\text{condition}] = 1 \) if the condition is true, otherwise it is 0. \( D_{\text{intra}} \) ranges from 0 to 1, with higher values indicating greater intra-class variation. As shown in Table \ref{tb9}, the intra-class diversity of synthetic dataset is 89.67\%.

\begin{figure}[t]
	\centering
	\includegraphics[scale=1,width=0.45\textwidth]{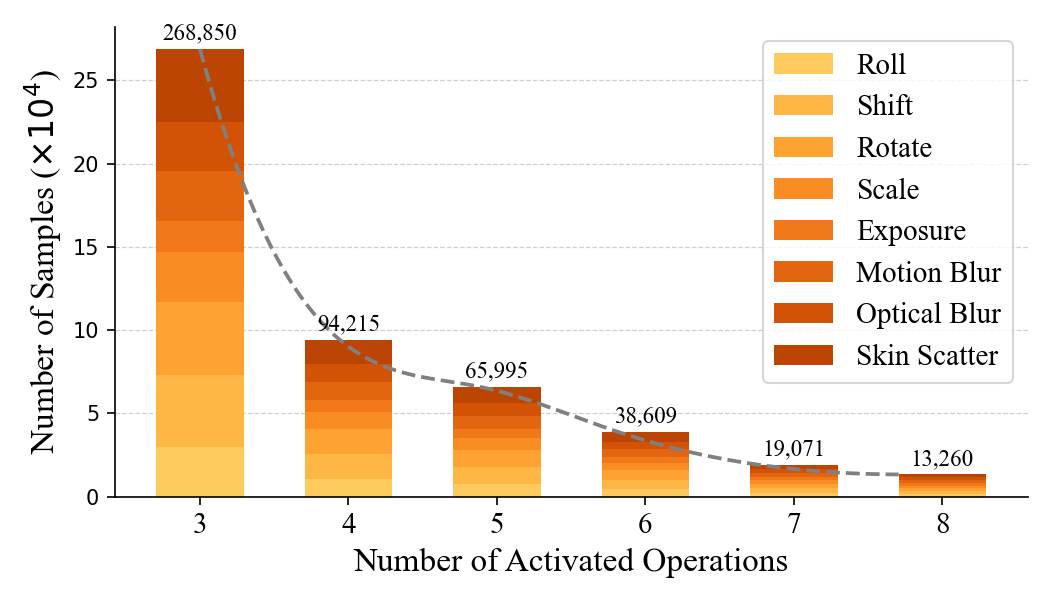}
	\caption{Distribution of the number of intra-class variations in 500,000 FVeinSyn generated samples.}
	\label{fig10}
\end{figure}

We plotted the matching score distributions of eight real finger vein datasets and the FVeinSyn generated synthetic dataset under varying numbers of identities and sample sizes, as shown in Fig.~\ref{fig9}. The results indicate that the synthetic and real datasets exhibit high consistency in key statistical measures, such as mean and variance, demonstrating significant statistical alignment. Specifically, the synthetic data closely matches the real data: the mean genuine/imposter matching scores of the real datasets range from $0.74–0.85 / 0.00–0.10$, while the synthetic dataset shows $0.84 / 0.03–0.04$, well within the typical range of real data; for variance, the real datasets’ genuine/imposter scores span $0.09–0.15 / 0.09–0.11$, and the synthetic data yields $0.07 / 0.08$, again demonstrating strong alignment with real distributions. This further confirms the reliability of FVeinSyn generated synthetic data in terms of both realism and diversity.

\begin{table*}[t]
	\centering
	\caption{Impact of different intra-class variations on recognition performance across datasets (TAR@FAR=$10^{-3}$)$\uparrow$.}
	\label{tab:intra_class_variation_mean}
	\renewcommand{\arraystretch}{0.82}
	
	\resizebox{\textwidth}{!}{
		\begin{tabular}{l*{10}{S[table-format=1.4]}}
			\toprule
			Intra-Class Variation 
			& {SDUMLA-FV} & {UTFVP} & {MMCBNU\_60000} & {PLUS-FV3} 
			& {FV-USM} & {HKPU-FV} & {SCUT-FV} & {THU-FV3} & {Avg.} \\
			\midrule
			
			Real data 
			& 0.4602 & 0.7593 & 0.8385 & 0.5385 & 0.7709 & 0.6447 & 0.5504 & 0.5799 & 0.6428 \\
			
			FVeinSyn (Mixed) 
			& 0.9402 & 0.9778 & 0.9538 & 0.9132 & 0.9551 & 0.9655 & 0.7078 & 0.9233 & 0.9171 \\
			
			\midrule
			\multicolumn{10}{l}{\textbf{Geometric Perturbations}} \\
			\midrule
			
			Shift 
			& 0.8994 & 0.9528 & 0.9395 & 0.8749 & 0.9009 & 0.9264 & 0.6642 & 0.8736 &{\textbf{0.8789}} \\
			
			Rotation 
			& 0.9283 & 0.9528 & 0.9209 & 0.8086 & 0.8875 & 0.8850 & 0.6965 & 0.8371 & $\underline{0.8646}$ \\
			
			Rolling 
			& 0.9107 & 0.9509 & 0.8901 & 0.8619 & 0.8726 & 0.8545 & 0.6675 & 0.7963 & 0.8506 \\
			
			Scaling 
			& 0.9034 & 0.9463 & 0.8993 & 0.8628 & 0.8945 & 0.8971 & 0.5996 & 0.8326 & 0.8545 \\
			
			\midrule
			\multicolumn{10}{l}{\textbf{Optical Degradation}} \\
			\midrule
			
			Exposure 
			& 0.8805 & 0.9194 & 0.8510 & 0.7863 & 0.9012 & 0.8710 & 0.6498 & 0.7952 & 0.8318 \\
			
			Skin Scattering 
			& 0.9090 & 0.9546 & 0.9447 & 0.8791 & 0.9046 & 0.8965 & 0.6635 & 0.8885 &\textbf{0.8801} \\
			
			Motion Blur 
			& 0.8868 & 0.9435 & 0.8873 & 0.8400 & 0.8870 & 0.8757 & 0.6249 & 0.8409 & 0.8483 \\
			
			Optical Blur 
			& 0.9149 & 0.9306 & 0.8833 & 0.8209 & 0.8977 & 0.8647 & 0.6498 & 0.8389 & $\underline{0.8501}$ \\
			
			\bottomrule
		\end{tabular}
	}
\end{table*}

\subsection{Ablation Study on Intra-Class Diversity Generator}
We conduct an ablation study to examine the contribution of different intra-class variation operators in the diversity generator. Specifically, we construct synthetic pretraining datasets with isolated transformations, including geometric perturbations and optical degradations, and pretrain separate models for evaluation under the same setting as open-set recognition.

As shown in Table~\ref{tab:intra_class_variation_mean}, all individual transformations improve recognition performance over the real-data baseline, indicating that both geometric and optical variations provide useful supervision during pretraining. These two categories contribute in a complementary manner, with geometric perturbations introducing structural variability and optical degradations modeling appearance-level changes.

The operator weights used in the diversity generator are consistent with the relative contributions observed in this ablation study. Specifically, shift, rotation, and skin scattering are assigned higher weights, while exposure is assigned a lower weight. The superior performance of the mixed setting further supports the use of empirically weighting for jointly modeling structural and appearance-level variations.

Furthermore, we analyze the distribution of intra-class variations in the generated samples. As shown in Fig.~\ref{fig10}, the number of variations per sample follows a long-tail distribution, where most samples contain only a few variations while samples with more variations gradually decrease. This trend is consistent with real-world data, indicating that the proposed diversity generator not only improves recognition performance but also produces samples with realistic statistical properties.

\subsection{Demographic Bias Analysis}
The training of the renderer depends on real data, the generated data may also inherit or amplify potential demographic biases in real data. This issue is particularly important for synthetic finger-vein data used as a pretraining foundation, because once unrecognized bias exists in the generation process, it may further affect the generalization performance of downstream recognition models across different demographic groups. Therefore, it is necessary to analyze the impact of pretraining with data synthesized by FVeinSyn on the recognition performance of different demographic groups.

Specifically, UTFVP and MMCBNU\_6000, two evaluation datasets with available demographic metadata, are used for evaluation. Under the same open-set recognition setting as the main experiments, test identities are grouped according to different demographic attributes for subgroup evaluation. For UTFVP, the recognition results are reported for gender and age groups. For MMCBNU\_6000, gender, age, race, and blood type are further analyzed. Table \ref{tab:demographic_analysis} reports the overall performance with and without FVeinSyn pretraining, as well as the recognition results of each demographic subgroup after synthetic pretraining, which are used to compare the differences between in-group performance and overall performance.

From the results of each demographic subgroup, it can be observed that the performance of different gender, age, race, and blood type groups is generally close to the overall performance, and the differences between subgroup performance and overall performance remain small. No specific group shows a persistent and significant deviation from the overall trend. At the same time, the performance gains brought by FVeinSyn pretraining are relatively stable across different groups, without being clearly concentrated on any particular group. No significant disproportionate performance degradation is observed for underrepresented groups.
\begin{table*}[t]
	\centering
	\caption{Demographic Fairness Analysis of Recognition Performance with Synthetic Data (TAR@FAR=$10^{-3}$)$\uparrow$.}
	\label{tab:demographic_analysis}
	\renewcommand{\arraystretch}{0.5}
	
	\resizebox{\textwidth}{!}{
		\begin{tabular}{
				c
				*{14}{S[table-format=1.4]}
			}
			\toprule
			\multirow{2}{*}{Dataset} 
			& \multicolumn{2}{c}{PT w/ Syn}
			& \multicolumn{2}{c}{Gender} 
			& \multicolumn{3}{c}{Age} 
			& \multicolumn{3}{c}{Race} 
			& \multicolumn{4}{c}{Blood Type} \\
			\cmidrule(lr){2-3} \cmidrule(lr){4-5} \cmidrule(lr){6-8} \cmidrule(lr){9-11} \cmidrule(lr){12-15}
			& {\XSolidBrush} & {\CheckmarkBold}
			& {Male} & {Female}
			& {$\leq$20} & {21--40} & {$\geq$41}
			& {Asian} & {Caucasian} & {African}
			& {A} & {B} & {AB} & {O} \\
			\midrule
			UTFVP 
			& 0.7593 & 0.9778
			& 0.9822 & 0.9222
			& 0.9815 & 0.9759 & 0.9514
			& {-} & {-} & {-}
			& {-} & {-} & {-} & {-} \\
			
			MMCBNU\_6000 
			& 0.8385 & 0.9538
			& 0.9458 & 0.9876
			& 1.0000 & 0.9494 & 0.9605
			& 0.9542 & 0.9531 & 0.9741
			& 0.9407 & 0.9874 & 0.9704 & 0.9663 \\
			\bottomrule
		\end{tabular}
	}
\end{table*}

\subsection{Computational Efficiency}
FVeinSyn adopts a three-stage generation pipeline, and the total time required to synthesize a finger vein image is the sum of the times consumed by each stage, as summarized in Table~\ref{tb10}. Specifically, the process consists of three steps: i) Identity Generation (285.6 ms), ii) Render (7.9 ms), and iii) Diversity Generation (0.7 ms). Consequently, the total time to generate a single synthetic sample is approximately 294.2 ms.

Each generated sample includes a 300 × 600 8-bit grayscale finger vein image, which requires about 55 KB ,storage in lossless Portable Network Graphics (PNG) format, an associated XML annotation file (approximately 2 KB), and a pixel-wise annotation mask representing the finger shape, vein pattern, and joint cavity regions (approximately 4 KB). FVeinSyn supports offline generation of synthetic finger vein samples for training, as well as on-the-fly generation during training, thereby effectively reducing the storage overhead.
\begin{table}[t]
	\centering
	\renewcommand{\arraystretch}{0.5}
	
	\setlength{\tabcolsep}{3pt} 
	\caption{Time consumption of Finger Vein Image Sample using FVeinSyn}
	\label{tb10}
	\begin{tabular}{l S[table-format=3.1] S[table-format=2.1] S[table-format=1.1] S[table-format=3.1]}
		\toprule
		Step & {Identity Generation} & {Render} & {Diversity Generation} & {Total} \\
		\midrule
		Time (ms) & 285.6 & 7.9 & 0.7 & 294.2 \\
		\bottomrule
	\end{tabular}
\end{table}

\section{Conclusion}
In this work, we presented FVeinSyn, a large-scale controllable synthetic data generation framework for finger vein recognition. By explicitly decoupling vascular topology synthesis from imaging appearance rendering, FVeinSyn effectively addresses the limitations caused by insufficient training data, such as low identity diversity and limited image realism. Our framework incorporates three collaborative modules: a finger vein identity generator to produce anatomically valid and identity-distinctive vein patterns, a cascaded region-aware GAN for realistic near-infrared rendering, and an intra-class diversity generator to simulate realistic intra-class variations. We used FVeinSyn to generate a dataset of 500,000 images across 10,000 vein identities and conducted extensive evaluations. Experimental results demonstrated that models trained with FVeinSyn achieve superior recognition performance, enhanced cross-domain robustness, and improved performance in data-limited scenarios, with an average recognition performance gain of 27.43\% compared to real-data baselines. These results confirm that FVeinSyn not only produces highly realistic and diverse synthetic samples but also significantly benefits downstream finger vein recognition tasks.

	\bibliographystyle{IEEEtran}
\bibliography{ref}

@article{sun2020face,
	title={Face spoofing detection based on local ternary label supervision in fully convolutional networks},
	author={Sun, Wenyun and Song, Yu and Chen, Changsheng and Huang, Jiwu and Kot, Alex C},
	journal={IEEE Trans. Inf. Forensics Secur.},
	volume={15},
	pages={3181--3196},
	year={2020},
	publisher={IEEE}
}

@article{cai2025rehearsal,
	title={Rehearsal-free and efficient continual learning for cross-domain face anti-spoofing},
	author={Cai, Rizhao and Cui, Yawen and Yu, Zitong and Lin, Xun and Chen, Changsheng and Kot, Alex},
	journal={IEEE Trans. Pattern Anal. Mach. Intell.},
	year={2025},
	publisher={IEEE}
}

@article{mittal2012making,
	title={Making a “completely blind” image quality analyzer},
	author={Mittal, Anish and Soundararajan, Rajiv and Bovik, Alan C},
	journal={IEEE Signal Process Lett.},
	volume={20},
	number={3},
	pages={209--212},
	year={2012},
	publisher={IEEE}
}

@article{moorthy2011blind,
	title={Blind image quality assessment: From natural scene statistics to perceptual quality},
	author={Moorthy, Anush Krishna and Bovik, Alan Conrad},
	journal={IEEE Trans. Image Process.},
	volume={20},
	number={12},
	pages={3350--3364},
	year={2011},
	publisher={IEEE}
}

@inproceedings{matkovic2005global,
	title={Global contrast factor-a new approach to image contrast.},
	author={Matkovic, Kresimir and Neumann, L{\'a}szl{\'o} and Neumann, Attila and Psik, Thomas and Purgathofer, Werner and others},
	booktitle={CAe},
	pages={159--167},
	year={2005}
}

@article{tsai2008information,
	title={Information entropy measure for evaluation of image quality},
	author={Tsai, Du-Yih and Lee, Yongbum and Matsuyama, Eri},
	journal={J. Digital Imaging},
	volume={21},
	number={3},
	pages={338--347},
	year={2008},
	publisher={Springer}
}

@article{zamir1976optimality,
	title={Optimality principles in arterial branching},
	author={Zamir, M},
	journal={J. Theor. Biol.},
	volume={62},
	number={1},
	pages={227--251},
	year={1976},
	publisher={Elsevier}
}

@article{murray1926physiological,
	title={The physiological principle of minimum work: I. The vascular system and the cost of blood volume},
	author={Murray, Cecil D},
	journal={PNAS},
	volume={12},
	number={3},
	pages={207--214},
	year={1926}
}

@article{qin2022local,
	title={Local attention transformer-based full-view finger-vein identification},
	author={Qin, Huafeng and Hu, Rongshan and El-Yacoubi, Mounim A and Li, Yantao and Gao, Xinbo},
	journal={IEEE Trans. Circuits Syst. Video Technol.},
	volume={33},
	number={6},
	pages={2767--2782},
	year={2022},
	publisher={IEEE}
}

@article{huang2024mirror,
	title={Mirror-Based Full-View Finger Vein Authentication With Illumination Adaptation},
	author={Huang, Junduan and Li, Zifeng and Bhattacharjee, Sushil and Marcel, S{\'e}bastien and Kang, Wenxiong},
	journal={IEEE Trans. Circuits Syst. Video Technol.},
	year={2024},
	publisher={IEEE}
}

@article{qin2024attention,
	title={Attention blstm-based temporal-spatial vein transformer for multi-view finger-vein recognition},
	author={Qin, Huafeng and Xiong, Zhipeng and Li, Yantao and El-Yacoubi, Mounim A and Wang, Jun},
	journal={IEEE Trans. Inf. Forensics Secur.},
	year={2024},
	publisher={IEEE}
}

@article{qin2021multi,
	title={Multi-scale and multi-direction GAN for CNN-based single palm-vein identification},
	author={Qin, Huafeng and El-Yacoubi, Mounim A and Li, Yantao and Liu, Chongwen},
	journal={IEEE Trans. Inf. Forensics Secur.},
	volume={16},
	pages={2652--2666},
	year={2021},
	publisher={IEEE}
}

@article{pan2020multi,
	title={Multi-scale deep representation aggregation for vein recognition},
	author={Pan, Zaiyu and Wang, Jun and Wang, Guoqing and Zhu, Jihong},
	journal={IEEE Trans. Inf. Forensics Secur.},
	volume={16},
	pages={1--15},
	year={2020},
	publisher={IEEE}
}

@article{shen2021finger,
	title={Finger vein recognition algorithm based on lightweight deep convolutional neural network},
	author={Shen, Jiaquan and Liu, Ningzhong and Xu, Chenglu and Sun, Han and Xiao, Yushun and Li, Deguang and Zhang, Yongxin},
	journal={IEEE Trans. Instrum. Meas.},
	volume={71},
	pages={1--13},
	year={2021},
	publisher={IEEE}
}

@article{ma2023focal,
	title={Focal contrastive learning for palm vein authentication},
	author={Ma, Yuer and Huang, Haizhen and Luo, Dacan and Zhang, Shifeng and Kang, Wenxiong and Xie, Di},
	journal={IEEE Trans. Instrum. Meas.},
	volume={72},
	pages={1--15},
	year={2023},
	publisher={IEEE}
}

@article{ou2024gscl,
	title={GSCL: Generative self-supervised contrastive learning for vein-based biometric verification},
	author={Ou, Wei-Feng and Po, Lai-Man and Huang, Xiu-Feng and Yu, Wing-Yin and Zhao, Yu-Zhi},
	journal={IEEE Trans. Biom. Behav. Identity Sci.},
	volume={6},
	number={2},
	pages={230--244},
	year={2024},
	publisher={IEEE}
}

@inproceedings{wang2025multi,
	title={A Multi-illumination Dataset and an Illumination Domain Adaptation Network for Finger Vein Identification},
	author={Wang, Huabin and Cheng, Yingfan and Zheng, Wu and Cheng, Jiayuan and Li, Xin and Li, Min and Liu, Fei},
	booktitle={MM - Proc. ACM Int. Conf. Multimed.},
	pages={7940--7948},
	year={2025}
}

@article{radzi2016finger,
	title={Finger-vein biometric identification using convolutional neural network},
	author={Radzi, Syafeeza Ahmad and Hani, Mohamed Khalil and Bakhteri, Rabia},
	journal={Turk. J. Electr. Eng. Comput. Sci.},
	volume={24},
	number={3},
	pages={1863--1878},
	year={2016}
}

@article{RLT,
	title={Feature extraction of finger-vein patterns based on repeated line tracking and its application to personal identification},
	author={Miura, Naoto and Nagasaka, Akio and Miyatake, Takafumi},
	journal={Mach. Vision Appl.},
	volume={15},
	number={4},
	pages={194--203},
	year={2004},
	publisher={Springer}
}

@article{MC,
	title={Extraction of finger-vein patterns using maximum curvature points in image profiles},
	author={Miura, Naoto and Nagasaka, Akio and Miyatake, Takafumi},
	journal={IEICE Trans. Inf. Syst.},
	volume={90},
	number={8},
	pages={1185--1194},
	year={2007},
	publisher={The Institute of Electronics, Information and Communication Engineers}
}

@article{IUWT,
	title={The undecimated wavelet decomposition and its reconstruction},
	author={Starck, Jean-Luc and Fadili, Jalal and Murtagh, Fionn},
	journal={IEEE Trans. Image Process.},
	volume={16},
	number={2},
	pages={297--309},
	year={2007},
	publisher={IEEE}
}

@inproceedings{WLD,
	title={Finger-vein authentication based on wide line detector and pattern normalization},
	author={Huang, Beining and Dai, Yanggang and Li, Rongfeng and Tang, Darun and Li, Wenxin},
	booktitle={Proc. Int. Conf. Pattern Recognit.},
	pages={1269--1272},
	year={2010},
	organization={IEEE}
}

@inproceedings{10.1145/3746027.3758253,
	author = {Wang, Yifan and Gui, Jie and Yu, Baosheng and Li, Qi and Sun, Zhenan and Kannala, Juho and Zhao, Guoying},
	title = {FingerVeinSyn-5M: A Million-Scale Dataset and Benchmark for Finger Vein Recognition},
	year = {2025},
	isbn = {9798400720352},
	publisher = {ACM},
	address = {New York, NY, USA},
	booktitle = {MM - Proc. ACM Int. Conf. Multimed.},
	pages = {13038–13045},
	numpages = {8},
	location = {Dublin, Ireland},
}

@article{casia-webface,
	title={Learning face representation from scratch},
	author={Yi, Dong and Lei, Zhen and Liao, Shengcai and Li, Stan Z},
	journal={arXiv preprint arXiv:1411.7923},
	year={2014}
}

@inproceedings{Guo2016MSCeleb1MAD,
	title={Ms-celeb-1m: A dataset and benchmark for large-scale face recognition},
	author={Guo, Yandong and Zhang, Lei and Hu, Yuxiao and He, Xiaodong and Gao, Jianfeng},
	booktitle={Eur. Conf. Comput. Vision, ECCV},
	pages={87--102},
	year={2016},
	organization={Springer}
}

@inproceedings{cao2018vggface2,
	title={Vggface2: A dataset for recognising faces across pose and age},
	author={Cao, Qiong and Shen, Li and Xie, Weidi and Parkhi, Omkar M and Zisserman, Andrew},
	booktitle={Proc. - IEEE Int. Conf. Autom. Face Gesture Recognit., FG},
	pages={67--74},
	year={2018},
	organization={IEEE}
}

@inproceedings{bansal2017umdfaces,
	title={UMDFaces: An Annotated Face Dataset for Training Deep Networks},
	author={Bansal, Ankan and Nanduri, Abhijit and Castillo, Carlos D. and Ranjan, Rajeev and Chellappa, Rama},
	booktitle={IEEE Int. Jt. Conf. Biom., IJCB},
	pages={464--473},
	year={2017},
	organization={IEEE}
}

@inproceedings{FaceScrub,
	title={A data-driven approach to cleaning large face datasets},
	author={Ng, Hong-Wei and Winkler, Stefan},
	booktitle={IEEE Int. Conf. Image Process., ICIP},
	pages={343--347},
	year={2014},
	organization={IEEE}
}

@article{wang2018devil,
	title={The Devil of Face Recognition is in the Noise},
	author={Wang, Fei and Chen, Liren and Li, Cheng and Huang, Shiyao and Chen, Yanjie and Qian, Chen and Loy, Chen Change},
	journal={arXiv preprint arXiv:1807.11649},
	year={2018}
}

@ARTICLE{9763004,
	author={Zhu, Zheng and Huang, Guan and Deng, Jiankang and Ye, Yun and Huang, Junjie and Chen, Xinze and Zhu, Jiagang and Yang, Tian and Du, Dalong and Lu, Jiwen and Zhou, Jie},
	journal={IEEE Trans. Pattern Anal. Mach. Intell.}, 
	title={WebFace260M: A Benchmark for Million-Scale Deep Face Recognition}, 
	year={2023},
	volume={45},
	number={2},
	pages={2627-2644},
	doi={10.1109/TPAMI.2022.3169734}}

@inproceedings{maio2004fvc2004,
	title={FVC2004: Third Fingerprint Verification Competition},
	author={Maio, Dario and Maltoni, Davide and Cappelli, Raffaele and Wayman, James L. and Jain, Anil K.},
	booktitle={Proc. Int. Conf. Biometric Authentication, ICBA},
	pages={1--7},
	year={2004},
	organization={Springer}
}

@inproceedings{arakala2013msp,
	title={MSP Fingerprint Matching Competition},
	author={Arakala, A. and Arora, J. and Jeffers, S. D. and Horadam, K. J.},
	booktitle={Proc. - Int. Conf. Biom., ICB},
	pages={1--8},
	year={2013},
	organization={IEEE}
}

@misc{nist302,
	title = {NIST Special Database 302},
	author = {{National Institute of Standards and Technology}},
	year = {2025},
	howpublished = {\url{https://www.nist.gov/itl/iad/image-group/nist-special-database-302}},
	note = {Accessed: 2025-10-29}
}

@misc{casiafingerprintv5,
	title = {CASIA-FingerprintV5 Dataset},
	author = {Chinese Academy of Sciences. Institute of Automation. National Laboratory of Pattern Recognition},
	howpublished = {\url{http://english.ia.cas.cn/db/201611/t20161101_169922.html}},
	note = {Accessed: 2025-10-29}
}

@article{fiumara2018nist,
	title={NIST special database 300: uncompressed plain and rolled images from fingerprint cards},
	author={Fiumara, Gregory P and Flanagan, Patricia A and Grantham, John D and Bandini, Bruce and Ko, Kenneth and Libert, John M},
	year={2018},
}

@inproceedings{yin2011sdumla,
	title={SDUMLA-HMT: A multimodal biometric database},
	author={Yin, Yilong and Liu, Lili and Sun, Xiwei},
	booktitle={Chinese Conf. Biometric Recognit., CCBR},
	pages={260--268},
	year={2011},
	organization={Springer}
}

@inproceedings{Ton_UTFVPDB2013,
	author = {B. T. {Ton} and R. N. J. {Veldhuis}},
	title = {A high quality finger vascular pattern dataset collected using a custom designed capturing device},
	booktitle = {Proc. - Int. Conf. Biom., ICB},
	location = {Madrid, Spain},
	pages = {1--5},
	year = {2013}
}

@inproceedings{lu2013available,
	title={An available database for the research of finger vein recognition},
	author={Lu, Yu and Xie, Shan Juan and Yoon, Sook and Wang, Zhihui and Park, Dong Sun},
	booktitle={Proc. Int. Congr. Image Signal Process., CISP},
	volume={1},
	pages={410--415},
	year={2013},
	organization={IEEE}
}

@INPROCEEDINGS{8698588,
	author={Kauba, Christof and Prommegger, Bernhard and Uhl, Andreas},
	booktitle={IEEE Int. Conf. Biom. Theory, Appl. Syst., BTAS}, 
	title={Focussing the Beam - A New Laser Illumination Based Data Set Providing Insights to Finger-Vein Recognition}, 
	year={2018},
	volume={},
	number={},
	pages={1-9},
	doi={10.1109/BTAS.2018.8698588}}

@article{asaari2014fusion,
	title={Fusion of band limited phase only correlation and width centroid contour distance for finger based biometrics},
	author={Asaari, Mohd Shahrimie Mohd and Suandi, Shahrel A and Rosdi, Bakhtiar Affendi},
	journal={Expert Syst. Appl.},
	volume={41},
	number={7},
	pages={3367--3382},
	year={2014},
	publisher={Elsevier}
}

@article{kumar2011human,
	title={Human identification using finger images},
	author={Kumar, Ajay and Zhou, Yingbo},
	journal={IEEE Trans. Image Process.},
	volume={21},
	number={4},
	pages={2228--2244},
	year={2011},
	publisher={IEEE}
}

@article{tang2019finger,
	title={Finger vein verification using a Siamese CNN},
	author={Tang, Su and Zhou, Shan and Kang, Wenxiong and Wu, Qiuxia and Deng, Feiqi},
	journal={IET Biom.},
	volume={8},
	number={5},
	pages={306--315},
	year={2019},
	publisher={Wiley Online Library}
}

@INPROCEEDINGS{6467066,
	author={Yang, Wenming and Huang, Xiaola and Liao, Qingmin},
	booktitle={Proc. Int. Conf. Image Process. ICIP}, 
	title={Fusion of finger vein and finger dorsal texture for personal identification based on Comparative Competitive Coding}, 
	year={2012},
	volume={},
	number={},
	pages={1141-1144},
	doi={10.1109/ICIP.2012.6467066}}

@inproceedings{vanoni2014cross,
	title={Cross-database evaluation using an open finger vein sensor},
	author={Vanoni, Matthias and Tome, Pedro and El Shafey, Laurent and Marcel, S{\'e}bastien},
	booktitle={BIOMS - IEEE Workshop Biom. Meas. Syst. Secur. Med. Appl., Proc.},
	pages={30--35},
	year={2014},
	organization={IEEE}
}

@article{ren2022dataset,
	title={A dataset and benchmark for multimodal biometric recognition based on fingerprint and finger vein},
	author={Ren, Hengyi and Sun, Lijuan and Guo, Jian and Han, Chong},
	journal={IEEE Trans. Inf. Forensics Secur.},
	volume={17},
	pages={2030--2043},
	year={2022},
	publisher={IEEE}
}

@article{kim202550,
	title={50 Years of Automated Face Recognition},
	author={Kim, Minchul and Jain, Anil and Liu, Xiaoming},
	journal={arXiv preprint arXiv:2505.24247},
	year={2025}
}

@ARTICLE{9209125,
	author={Cavazos, Jacqueline G. and Phillips, P. Jonathon and Castillo, Carlos D. and O’Toole, Alice J.},
	journal={IEEE Trans. Biom. Behav. Identity Sci.}, 
	title={Accuracy Comparison Across Face Recognition Algorithms: Where Are We on Measuring Race Bias?}, 
	year={2021},
	volume={3},
	number={1},
	pages={101-111}
	}

@inproceedings{fan2025divtrackee,
	title     = {DivTrackee versus DynTracker: Promoting Diversity in Anti-Facial Recognition against Dynamic {FR} Strategy},
	author    = {Fan, Wenshu and Zhang, Minxing and Li, Hongwei and Jiang, Wenbo and Chen, Hanxiao and Yue, Xiangyu and Backes, Michael and Zhang, Xiao},
	booktitle = {CCS - Proc. ACM SIGSAC Conf. Comput. Commun. Secur.},
	year      = {2025},
}

@inproceedings{flotho2025t,
	title={T-FAKE: Synthesizing Thermal Images for Facial Landmarking},
	author={Flotho, Philipp and Piening, Moritz and Kukleva, Anna and Steidl, Gabriele},
	booktitle={Proc IEEE Int Conf Comput Vision, CVPR},
	pages={26356--26366},
	year={2025}
}

@ARTICLE{9737477,
	author={Yang, Ziming and Liang, Jian and Fu, Chaoyou and Luo, Mandi and Zhang, Xiao-Yu},
	journal={IEEE Trans. Inf. Forensics Secur.}, 
	title={Heterogeneous Face Recognition via Face Synthesis With Identity-Attribute Disentanglement}, 
	year={2022},
	volume={17},
	number={},
	pages={1344-1358}}

@article{yao2025synthetic,
	title={Synthetic Face Ageing: Evaluation, Analysis and Facilitation of Age-Robust Facial Recognition Algorithms},
	author={Yao, Wang and Farooq, Muhammad Ali and Lemley, Joseph and Corcoran, Peter},
	journal={IEEE Trans. Biom. Behav. Identity Sci.},
	year={2025},
	publisher={IEEE}
}

@article{grosz2024universal,
	title={Universal fingerprint generation: Controllable diffusion model with multimodal conditions},
	author={Grosz, Steven A and Jain, Anil K},
	journal={IEEE Trans. Pattern Anal. Mach. Intell.},
	year={2024},
	publisher={IEEE}
}

@article{dong2023synthesis,
	title={Synthesis of multi-view 3D fingerprints to advance contactless fingerprint identification},
	author={Dong, Chengdong and Kumar, Ajay},
	journal={IEEE Trans. Pattern Anal. Mach. Intell.},
	volume={45},
	number={11},
	pages={13134--13151},
	year={2023},
	publisher={IEEE}
}

@article{grosz2022spoofgan,
	title={Spoofgan: Synthetic fingerprint spoof images},
	author={Grosz, Steven A and Jain, Anil K},
	journal={IEEE Trans. Inf. Forensics Secur.},
	volume={18},
	pages={730--743},
	year={2022},
	publisher={IEEE}
}

@article{sun2023zjut,
	title={ZJUT-EIFD: A synchronously collected external and internal fingerprint database},
	author={Sun, Haohao and Wang, Haixia and Zhang, Yilong and Liang, Ronghua and Chen, Peng and Feng, Jianjiang},
	journal={IEEE Trans. Pattern Anal. Mach. Intell.},
	volume={46},
	number={4},
	pages={2267--2284},
	year={2023},
	publisher={IEEE}
}

@article{dong2025bridging,
	title={Bridging Dimensions in Fingerprints to Advance Distinctiveness: Recovering 3D Minutiae from a Single Contactless 2D Fingerprint Image},
	author={Dong, Chengdong and Kumar, Ajay},
	journal={IEEE Trans. Pattern Anal. Mach. Intell.},
	year={2025},
	publisher={IEEE}
}

@inproceedings{hillerstrom2014generating,
	title={Generating and analyzing synthetic finger vein images},
	author={Hillerstr{\"o}m, Fieke and Kumar, Ajay and Veldhuis, Raymond},
	booktitle={BIOSIG - Proc. Int. Conf. Biom. Spec. Interest Group},
	pages={1--9},
	year={2014},
	organization={IEEE}
}

@article{ou2022gan,
	title={GAN-based inter-class sample generation for contrastive learning of vein image representations},
	author={Ou, Wei-Feng and Po, Lai-Man and Zhou, Chang and Xian, Peng-Fei and Xiong, Jing-Jing},
	journal={IEEE Trans. Biom. Behav. Identity Sci.},
	volume={4},
	number={2},
	pages={249--262},
	year={2022},
	publisher={IEEE}
}

@ARTICLE{9893541,
	author={Engelsma, Joshua James and Grosz, Steven and Jain, Anil K.},
	journal={IEEE Trans. Pattern Anal. Mach. Intell.}, 
	title={PrintsGAN: Synthetic Fingerprint Generator}, 
	year={2023},
	volume={45},
	number={5},
	pages={6111-6124},
	doi={10.1109/TPAMI.2022.3204591}}

@ARTICLE{10620353,
	author={Wang, Yifan and Gui, Jie and Tang, Yuan Yan and Kwok, James Tin-Yau},
	journal={IEEE Trans. Inf. Forensics Secur.}, 
	title={CFVNet: An End-to-End Cancelable Finger Vein Network for Recognition}, 
	year={2024},
	volume={19},
	number={},
	pages={7810-7823},
	doi={10.1109/TIFS.2024.3436528}}

@article{WANG2023119874,
	title = {Residual Gabor convolutional network and FV-Mix exponential level data augmentation strategy for finger vein recognition},
	journal = {Expert Syst. Appl.},
	volume = {223},
	pages = {119874},
	year = {2023},
	issn = {0957-4174},
	doi = {https://doi.org/10.1016/j.eswa.2023.119874},
	url = {https://www.sciencedirect.com/science/article/pii/S0957417423003755},
	author = {Yifan Wang and Huimin Lu and Xiwen Qin and Jianwei Guo}}

@article{sweeney2024unsupervised,
	title={Unsupervised segmentation of 3D microvascular photoacoustic images using deep generative learning},
	author={Sweeney, Paul W and Hacker, Lina and Lefebvre, Thierry L and Brown, Emma L and Gr{\"o}hl, Janek and Bohndiek, Sarah E},
	journal={Adv. Sci.},
	volume={11},
	number={32},
	pages={2402195},
	year={2024},
	publisher={Wiley Online Library}
}

@inproceedings{shang2025pvtree,
	title={PVTree: Realistic and Controllable Palm Vein Generation for Recognition Tasks},
	author={Shang, Sheng and Zhao, Chenglong and Zhang, Ruixin and Jin, Jianlong and Zhang, Jingyun and Guo, Rizen and Ding, Shouhong and Wu, Yunsheng and Zhao, Yang and Jia, Wei},
	booktitle={Proc. AAAI Conf. Artif. Intell.},
	volume={39},
	number={7},
	pages={6767--6775},
	year={2025}
}

@inproceedings{jin2024pce,
	title={Pce-palm: Palm crease energy based two-stage realistic pseudo-palmprint generation},
	author={Jin, Jianlong and Shen, Lei and Zhang, Ruixin and Zhao, Chenglong and Jin, Ge and Zhang, Jingyun and Ding, Shouhong and Zhao, Yang and Jia, Wei},
	booktitle={Proc. AAAI Conf. Artif. Intell.},
	volume={38},
	number={3},
	pages={2616--2624},
	year={2024}
}

@InProceedings{Shen_2023_ICCV,
	author    = {Shen, Lei and Jin, Jianlong and Zhang, Ruixin and Li, Huaen and Zhao, Kai and Zhang, Yingyi and Zhang, Jingyun and Ding, Shouhong and Zhao, Yang and Jia, Wei},
	title     = {RPG-Palm: Realistic Pseudo-data Generation for Palmprint Recognition},
	booktitle = {Proc IEEE Int Conf Comput Vision, ICCV},
	month     = {October},
	year      = {2023},
	pages     = {19605-19616}
}

@inproceedings{zhao2022bezierpalm,
	title={B{\'e}zierpalm: A free lunch for palmprint recognition},
	author={Zhao, Kai and Shen, Lei and Zhang, Yingyi and Zhou, Chuhan and Wang, Tao and Zhang, Ruixin and Ding, Shouhong and Jia, Wei and Shen, Wei},
	booktitle={Eur. Conf. Comput. Vision, ECCV},
	pages={19--36},
	year={2022},
	organization={Springer}
}

@inproceedings{jin2025diffpalm,
	title={Diff-Palm: Realistic Palmprint Generation with Polynomial Creases and Intra-Class Variation Controllable Diffusion Models},
	author={Jin, Jianlong and Zhao, Chenglong and Zhang, Ruixin and Shang, Sheng and Xu, Jianqing and Zhang, Jingyun and Wang, ShaoMing and Zhao, Yang and Ding, Shouhong and Jia, Wei and others},
	booktitle={Proc IEEE Int Conf Comput Vision, CVPR},
	pages={26367--26376},
	year={2025}
}

@INPROCEEDINGS{10204758,
	author={Kim, Minchul and Liu, Feng and Jain, Anil and Liu, Xiaoming},
	booktitle={Proc IEEE Int Conf Comput Vision, CVPR}, 
	title={DCFace: Synthetic Face Generation with Dual Condition Diffusion Model}, 
	year={2023},
	volume={},
	number={},
	pages={12715-12725},
	doi={10.1109/CVPR52729.2023.01223}}

@ARTICLE{10399793,
	author={Melnik, Andrew and Miasayedzenkau, Maksim and Makaravets, Dzianis and Pirshtuk, Dzianis and Akbulut, Eren and Holzmann, Dennis and Renusch, Tarek and Reichert, Gustav and Ritter, Helge},
	journal={IEEE Trans. Pattern Anal. Mach. Intell.}, 
	title={Face Generation and Editing With StyleGAN: A Survey}, 
	year={2024},
	volume={46},
	number={5},
	pages={3557-3576}}

@InProceedings{Kapitanov_2024_WACV,
	author    = {Kapitanov, Alexander and Kvanchiani, Karina and Nagaev, Alexander and Kraynov, Roman and Makhliarchuk, Andrei},
	title     = {HaGRID -- HAnd Gesture Recognition Image Dataset},
	booktitle = {Proc. - IEEE/CVF Winter Conf. Appl. Comput. Vis., WACV},
	month     = {January},
	year      = {2024},
	pages     = {4572-4581}
}

@article{kirillov2023segany,
	title={Segment Anything},
	author={Kirillov, Alexander and Mintun, Eric and Ravi, Nikhila and Mao, Hanzi and Rolland, Chloe and Gustafson, Laura and Xiao, Tete and Whitehead, Spencer and Berg, Alexander C. and Lo, Wan-Yen and Doll{\'a}r, Piotr and Girshick, Ross},
	journal={arXiv:2304.02643},
	year={2023}
}

@ARTICLE{9363648,
	author={Hou, Borui and Yan, Ruqiang},
	journal={IEEE Trans. Instrum. Meas.}, 
	title={ArcVein-Arccosine Center Loss for Finger Vein Verification}, 
	year={2021},
	volume={70},
	number={},
	pages={1-11},
	doi={10.1109/TIM.2021.3062164}}

@ARTICLE{8395431,
	author={Das, Rig and Piciucco, Emanuela and Maiorana, Emanuele and Campisi, Patrizio},
	journal={IEEE Trans. Inf. Forensics Secur.}, 
	title={Convolutional Neural Network for Finger-Vein-Based Biometric Identification}, 
	year={2019},
	volume={14},
	number={2},
	pages={360-373},
	doi={10.1109/TIFS.2018.2850320}}

@ARTICLE{8979362,
	author={Kuzu, Ridvan Salih and Piciucco, Emanuela and Maiorana, Emanuele and Campisi, Patrizio},
	journal={IEEE Trans. Inf. Forensics Secur.}, 
	title={On-the-Fly Finger-Vein-Based Biometric Recognition Using Deep Neural Networks}, 
	year={2020},
	volume={15},
	number={},
	pages={2641-2654},
	doi={10.1109/TIFS.2020.2971144}}

@ARTICLE{9408606,
	author={Li, Shuyi and Zhang, Bob},
	journal={IEEE Trans. Inf. Forensics Secur.}, 
	title={Joint Discriminative Sparse Coding for Robust Hand-Based Multimodal Recognition}, 
	year={2021},
	volume={16},
	number={},
	pages={3186-3198},
	doi={10.1109/TIFS.2021.3074315}}

@article{ren2025iis,
	title={IIS-FVIQA: Finger Vein Image Quality Assessment with intra-class and inter-class similarity},
	author={Ren, Hengyi and Sun, Lijuan and Fan, Xijian and Cao, Ying and Ye, Qiaolin},
	journal={Pattern Recognit.},
	volume={158},
	pages={111056},
	year={2025},
	publisher={Elsevier}
}

@ARTICLE{9973284,
	author={Zhao, Pengyang and Chen, Zhiquan and Xue, Jing-Hao and Feng, Jianjiang and Yang, Wenming and Liao, Qingmin and Zhou, Jie},
	journal={IEEE Trans. Biom. Behav. Identity Sci.}, 
	title={Single-Sample Finger Vein Recognition via Competitive and Progressive Sparse Representation}, 
	year={2023},
	volume={5},
	number={2},
	pages={209-220},
	doi={10.1109/TBIOM.2022.3226270}}

@ARTICLE{10023509,
	author={Huang, Junduan and Zheng, An and Shakeel, M. Saad and Yang, Weili and Kang, Wenxiong},
	journal={IEEE Trans. Inf. Forensics Secur.}, 
	title={FVFSNet: Frequency-Spatial Coupling Network for Finger Vein Authentication}, 
	year={2023},
	volume={18},
	number={},
	pages={1322-1334},
	doi={10.1109/TIFS.2023.3238546}}

@ARTICLE{10530126,
	author={Li, Yantao and Tang, Congyi and Deng, Shaojiang and Qin, Huafeng and Huang, Hongyu},
	journal={IEEE Trans. Instrum. Meas.}, 
	title={Mixed Automatic Adversarial Augmentation Network for Finger-Vein Recognition}, 
	year={2024},
	volume={73},
	number={},
	pages={1-10},
	doi={10.1109/TIM.2024.3400355}}

@article{lindenmayer1968mathematical,
	title={Mathematical models for cellular interactions in development I. Filaments with one-sided inputs},
	author={Lindenmayer, Aristid},
	journal={J. Theor. Biol.},
	volume={18},
	number={3},
	pages={280--299},
	year={1968},
	publisher={Elsevier}
}

@book{prusinkiewicz2013lindenmayer,
	title={Lindenmayer systems, fractals, and plants},
	author={Prusinkiewicz, Przemyslaw and Hanan, James},
	volume={79},
	year={2013},
	publisher={Springer Science \& Business Media}
}

@article{wang2025colorvein,
	title={ColorVein: Colorful Cancelable Vein Biometrics},
	author={Wang, Yifan and Gui, Jie and Shi, Xinli and Gui, Linqing and Tang, Yuan Yan and Kwok, James Tin-Yau},
	journal={IEEE Trans. Inf. Forensics Secur.},
	year={2025},
	publisher={IEEE}
}

@inproceedings{zhu2017unpaired,
	title={Unpaired image-to-image translation using cycle-consistent adversarial networks},
	author={Zhu, Jun-Yan and Park, Taesung and Isola, Phillip and Efros, Alexei A},
	booktitle={Proc IEEE Int Conf Comput Vision, ICCV},
	pages={2223--2232},
	year={2017}
}

@inproceedings{deng2019arcface,
	title={Arcface: Additive angular margin loss for deep face recognition},
	author={Deng, Jiankang and Guo, Jia and Xue, Niannan and Zafeiriou, Stefanos},
	booktitle={Proc IEEE Int Conf Comput Vision, CVPR},
	year={2019}
}

@inproceedings{he2016deep,
	title={Deep residual learning for image recognition},
	author={He, Kaiming and Zhang, Xiangyu and Ren, Shaoqing and Sun, Jian},
	booktitle={Proc IEEE Int Conf Comput Vision, CVPR},
	pages={770--778},
	year={2016}
}

@inproceedings{nayar1999vision,
	title={Vision in bad weather},
	author={Nayar, Shree K and Narasimhan, Srinivasa G},
	booktitle={Proc IEEE Int Conf Comput Vision, ICCV},
	volume={2},
	pages={820--827},
	year={1999},
	organization={IEEE}
}

@inproceedings{10.1145/3664647.3680635,
	author = {Lin, Jingyu and Zhao, Guiqin and Xu, Jing and Wang, Guoli and Wang, Zejin and Dantcheva, Antitza and Du, Lan and Chen, Cunjian},
	title = {DiffTV: Identity-Preserved Thermal-to-Visible Face Translation via Feature Alignment and Dual-Stage Conditions},
	year = {2024},
	isbn = {9798400706868},
	publisher = {ACM},
	address = {New York, NY, USA},
	doi = {10.1145/3664647.3680635},
	booktitle = {MM - Proc. ACM Int. Conf. Multimed.},
	pages = {10930–10938},
	numpages = {9},
	location = {Melbourne VIC, Australia},
}

@inproceedings{10.1145/3664647.3680704,
	author = {Wang, Tao and Zhang, Yushu and Xiao, Xiangli and Yuan, Lin and Xia, Zhihua and Weng, Jian},
	title = {Make Privacy Renewable! Generating Privacy-Preserving Faces Supporting Cancelable Biometric Recognition},
	year = {2024},
	isbn = {9798400706868},
	publisher = {ACM},
	address = {New York, NY, USA},
	booktitle = {MM - Proc. ACM Int. Conf. Multimed.},
	pages = {10268–10276},
	numpages = {9},
	location = {Melbourne VIC, Australia},
}

\vfill

\end{document}